\documentclass[lettersize,journal]{IEEEtran}
\usepackage{amsmath,amsfonts}
\usepackage{algorithmic}
\usepackage{algorithm}
\usepackage{array}
\usepackage[caption=false,font=normalsize,labelfont=sf,textfont=sf]{subfig}
\usepackage{textcomp}
\usepackage{stfloats}
\usepackage{url}
\usepackage{verbatim}
\usepackage{graphicx}
\usepackage{cite}
\usepackage{amssymb}
\usepackage{bbding}
\usepackage{color}
\usepackage{xcolor}
\usepackage{colortbl}
\definecolor{ours}{gray}{.95}
\definecolor{indiagreen}{rgb}{0.07, 0.53, 0.03}
\definecolor{Revision}{RGB}{255,0,0}
\usepackage{multirow}
\usepackage{booktabs}

\def\ourmodel{EMIP}
\def\eg{\emph{e.g.}}

\def\ie{\emph{i.e.}}

\title{Explicit Motion Handling and Interactive Prompting for Video Camouflaged Object Detection}

\author{Xin Zhang,~Tao Xiao,~Ge-Peng Ji,~Xuan Wu,~Keren Fu,~and~Qijun Zhao

\IEEEcompsocitemizethanks{\IEEEcompsocthanksitem 
Manuscript received on June 1, 2024. 
\textit{(Corresponding author: Keren Fu.)}
\IEEEcompsocthanksitem Xin Zhang is with the National Key Laboratory of Fundamental Science on Synthetic Vision, Sichuan University, Chengdu 610065, China. (E-mail: zhangxinchina1314@gmail.com)
\IEEEcompsocthanksitem Tao Xiao, Xuan Wu, Keren Fu, and Qijun Zhao are with the College of Computer Science, and
the National Key Laboratory of Fundamental Science on Synthetic Vision,
Sichuan University, Chengdu 610065, China. (E-mail: 2021223045243@stu.scu.edu.cn; 2023223040230@stu.scu.edu.cn; fkrsuper@scu.edu.cn; qjzhao@scu.edu.cn)
\IEEEcompsocthanksitem Ge-Peng Ji is with the School of Computing, Australian National University, Canberra, Australia. (E-mail: gepengai.ji@gmail.com)

}
}

\begin{document}
\maketitle

\markboth{Journal of \LaTeX\ Class Files,~Vol.~14, No.~8, August~2021}%
{Shell \MakeLowercase{\textit{et al.}}: A Sample Article Using IEEEtran.cls for IEEE Journals}


\begin{abstract}
Camouflage poses notable challenges in distinguishing a static target, as it usually blends seamlessly with the background. However, any movement by the target can disrupt this disguise, making it detectable. Existing video camouflaged object detection (VCOD) approaches take noisy motion estimation as input or model motion implicitly, restricting detection performance in complex dynamic scenes.
In this paper, we propose a novel \textbf{E}xplicit \textbf{M}otion handling and \textbf{I}nteractive \textbf{P}rompting framework for VCOD, dubbed EMIP, which handles motion cues explicitly using a frozen pre-trained optical flow fundamental model. EMIP is characterized by a two-stream architecture for simultaneously conducting camouflaged segmentation and optical flow estimation. Interactions across the dual streams are realized in an interactive prompting way that is inspired by emerging visual prompt learning. Two learnable modules, i.e. the camouflaged feeder and motion collector, are designed to incorporate segmentation-to-motion and motion-to-segmentation prompts, respectively, and enhance outputs of the both streams. The prompt fed to the motion stream is learned by supervising optical flow in a self-supervised manner.
Furthermore, we show that long-term historical information can also be incorporated as a prompt into EMIP and achieve more robust results with temporal consistency.
By leveraging promoting techniques based on EMIP, the proposed long-term model EMIP$^\dag$ incurs lower training cost with only 8.5M trainable parameters (less than 8\% of the total model parameters).
Experimental results demonstrate that both EMIP and EMIP$^\dag$ set new state-of-the-art records on popular VCOD benchmarks. Additionally, comparative evaluations against other video segmentation models on a wider range of video segmentation tasks demonstrate the robustness and superior generalization capabilities of EMIP.
Our code is made publicly available at \url{https://github.com/zhangxin06/EMIP}.

\end{abstract}


\section{Introduction}
\IEEEPARstart{C}{amouflaged} object detection (COD) aims at detecting and segmenting those \textit{hidden objects} that exhibit high intrinsic similarity to their backgrounds. The inherent complexity of distinguishing camouflaged objects from their surroundings poses unique challenges when compared to general object detection \cite{zhao2019object} and salient object detection (SOD) \cite{zhuge2022salient}. Recently, it has attracted interest of many researchers and facilities broad applications, \eg, medical image segmentation~\cite{ji2022vps,wu2023medical} and industrial inspection~\cite{he2019fully,yang2022surface}. While significant progress has been made in image-based \cite{fan2021concealed,fan2020Camouflage,ji2023gradient} and video-based \cite{cheng2022implicit,lamdouar2020betrayed,xie2022segmenting} COD tasks, there still remains substantial room for development due to intrinsic difficulty of the tasks. 

\begin{figure}[t!]
  \centering
  \includegraphics[width=0.48\textwidth]{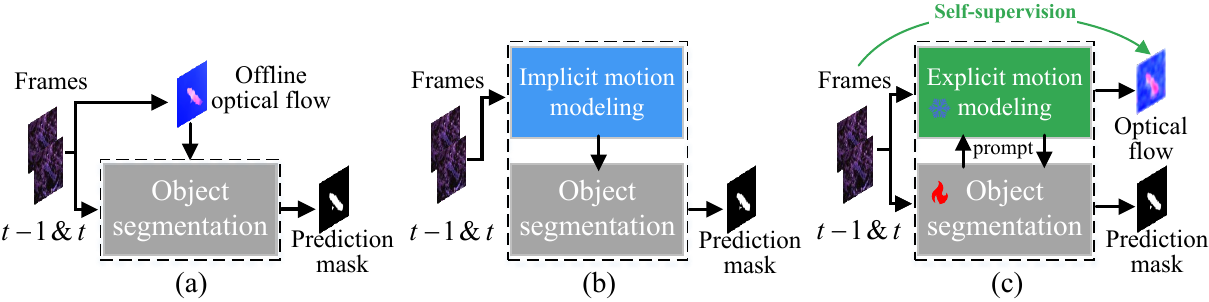}
  \caption{Different strategies of motion handling in VCOD: (a) Directly feed optical flow maps \protect\cite{lamdouar2020betrayed,yang2021selfsupervised}; (b) Learn implicit motion cues and subsequently utilize them for mask decoding \protect\cite{cheng2022implicit}; (c) The proposed interactive prompting paradigm 
  handles motion cues explicitly using a pre-trained optical flow model, and simultaneously conduct optical flow estimation and camouflaged object segmentation.
  The fire/snowflake symbols denote that most of the model parameters are learnable/frozen in the proposed scheme.}
  \label{fig1}
\end{figure}

Image-based COD approaches identify camouflaged targets using a static image. Typically, these targets exhibit strong visual resemblances to their backgrounds in terms of texture, color, or edges, posing a challenge for detection using appearance or geometric cues alone. 
Consequently, motion cues have been investigated in previous works~\cite{lamdouar2020betrayed,yang2021selfsupervised} for video camouflaged object detection (VCOD) task. 
Fig. \ref{fig1} summarizes the strategies adopted in handling motion cues by these previous works. As presented in Fig. \ref{fig1} (a), off-the-shelf motion estimators (\eg, \cite{teed2020raft,liu2020learning}) are directly employed for generating offline optical flow, serving as motion cues for identifying camouflaged objects.
Notably, offline motion estimation, particularly in camouflaged scenarios, poses significant challenges for common optical flow estimators, often leading to noisy and inaccurate output optical flow. Such erroneous input could misguide detectors and thereby hinder overall performance.
To address this issue, SLT-Net \cite{cheng2022implicit} proposes a distinct approach by implicit motion modeling in an online fashion (Fig. \ref{fig1} (b)). While the overall motion modeling part is learnable, learning reliable motion of camouflaged objects from limited VCOD data may be intractable compared to using extensive training data for training common optical flow estimators \cite{teed2020raft,xu2022gmflow}. Besides, due to the implicit nature of SLT-Net, no explicit regularization or evaluation can be adhered to the motion part, further making the learned motion less effective and reliable.
Also, we believe that beyond motion, appearance cues are an important factor conducive to detection as well, since objects could be motion-free or still.

In this paper, we propose a novel Explicit Motion handling and Interactive Prompting framework for VCOD (Fig. \ref{fig1} (c)), dubbed \ourmodel, which handles motion cues explicitly by freezing the upstream optical flow model mostly. 
EMIP is characterized by a two-stream architecture for simultaneously addressing camouflaged segmentation and optical flow estimation. 
Inspired by emerging advances of the visual prompt learning~\cite{sandler2022fine,jia2022visual,zhu2023visual}, we design an interactive prompting scheme to achieve interactions across the dual streams, as shown in Fig. \ref{fig1} (c).
Two modules, namely the camouflage feeder and motion collector, are designed to incorporate segmentation-to-motion and motion-to-segmentation prompts, respectively, and enhance outputs of the dual streams. 
Due to the absence of authentic optical flow for current VCOD datasets, we learn the prompt fed to the motion stream by supervising optical flow in a self-supervised manner. We also propose a long-term variant of EMIP by formulating historical features into the prompt to mitigate short-term prediction errors and further improve the accuracy. Benefiting from the above elaborate designs, the proposed \ourmodel~ effectively leverages noise-robust motion to detect and segment those video camouflaged objects. The main contributions are summarized below:
\begin{itemize}
    \item We propose a novel framework for VCOD, dubbed EMIP, which handles motion cues explicitly using a frozen pre-trained optical flow fundamental model. EMIP is formulated into a novel two-stream architecture for simultaneously conducting camouflaged segmentation and optical flow estimation.
    
    \item Inspired by visual prompt learning, the interactions across the two streams are realized in an interactive prompting way, and we propose two modules, \emph{i.e.}, the camouflaged feeder and motion collector, to incorporate segmentation-to-motion and motion-to-segmentation prompts, respectively.
    
    \item We also propose a long-term variant of \ourmodel, dubbed EMIP$^\dag$, by formulating historical features into the prompt to mitigate short-term prediction errors.
    
    \item \ourmodel~together with its long-term variant \ourmodel$^\dag$, achieve new state-of-the-art records and outperform previous models by notable margins ($\sim$17.0\%/5.5\% average improvement from \ourmodel~on F-measure/S-measure over the previous cutting-edge model SLT-Net).
\end{itemize}

The remainder of the paper is organized as follows: Section II discusses related work on image and video-based COD, closely related video object segmentation, salient object detection, and visual prompt learning. Section III describes the proposed models in detail. Section IV includes experimental results, comparisons, and analyses. Finally, conclusions are drawn in Section V.

\section{Related Work}
\subsection{Image-based COD}
Methods in this type aim to discern camouflaged objects from a single RGB image. 
Early COD methods~\cite{pan2011study,liu2012foreground} relied on hand-crafted features to find targets hidden in the background. 
Li~\textit{et al.}~\cite{Li2017TGWV} proposed a texture guided weighted voting strategy to detect camouflaged objects. Then they further introduced a fusion framework~\cite{Li2018fusion} to address camouflaged problems in the wavelet domain.
And more, Garcia~\cite{garcia2020background} review the background subtraction methods and outlook future direction. 
However, with the advent of deep learning, COD methods have undergone substantial advancement in recent years. Inspired by natural predatory behavior, SINet-V2~\cite{fan2021concealed} and PFNet ~\cite{mei2021camouflaged} employed a coarse-to-fine strategy. They first generated a preliminary location map for camouflaged objects and then refined it for segmentation. To enhance performance, several studies integrated auxiliary task into a joint learning framework. MGL~\cite{zhai2021mutual} combined classification or boundary detection task with COD.
Liu~\textit{et al.}~\cite{liu2021POCINet} investigated the part-object relationship to discover camouflaged objects.
ZoomNet~\cite{pang2022zoom} employed a zooming in and out strategy to the original inputs and processed appearance features at three different scales.
Then Pang~\textit{et al.} extended ZoomNet to ZoomNeXt~\cite{pang2024zoomnext}, which simultaneously addresses image and video camouflaged object detection.
Jia~\textit{et al.}~\cite{jia2022segmar} proposed the SegMaR framework, an iterative refinement approach designed to locate, magnify, and detect camouflaged objects.
Ji~\textit{et al.}~\cite{ji2023gradient} designed a two-branch framework to encode the context and texture of camouflage objects under gradient supervision.
Huang~\textit{et al.}~\cite{huang2023feature} designed progressively neighboring token enhanced decoder to exploit imperceptible cues for detecting camouflaged objects.
Zhang~\textit{et al.}~\cite{zhang2023predictive} introduced predictive uncertainty estimation framework to address model and data uncertainty simultaneously in camouflaged scenes .
HitNet~\cite{hu2023high} elevated low-resolution representations by leveraging high-resolution features in an iterative feedback loop, effectively mitigating challenges such as edge blurring and detail degradation.
Methods~\cite{he2023FEDER,cong2023frequency} designed frameworks to mine the subtle cues of camouflaged objects in the frequency domain.
Yao~\textit{et al.}~\cite{Yao2024HGINet} proposed a hierarchical graph interaction network to refine ambiguous regions.
Hao~\textit{et al.}~\cite{hao2025senet} proposed a simple yet effective general architecture for both COD and SOD tasks.

\subsection{Video-based COD}
For the VCOD task, motion cues are crucial to camouflaged object detection. 
Bidau~\textit{et al.} \cite{bideau2016s} introduced a method by approximating various motion models derived from dense optical flow.
Zhang~\textit{et al.}~\cite{zhang2017CM} proposed a camouflage modeling strategy and fused it with discriminative modeling in a Bayesian framework for moving object detection.
Lamdouar~\textit{et al.}\cite{lamdouar2020betrayed} introduced a video registration and segmentation network to detect camouflaged objects, employing optical flow and a difference image as inputs. However, the utilization of inaccurate optical flow may result in accumulative errors in mask prediction.
To address this challenge, Cheng~\textit{et al.}~\cite{cheng2022implicit} proposed a two-stage model that implicitly models and leverages motion information. 
Subsequently, to eliminate inaccuracies stemming from implicit motion modeling in SLT-Net~\cite{cheng2022implicit}, Hui~\textit{et al.}~\cite{hui2024implict-explicit} introduced a motion-induced consistency preserving approach between frames with a feature pyramid framework.
Lu~\textit{et al.}~\cite{lu2025weakly-supervised} introduced a weakly-supervised framework for VCOD.
Hui \textit{et al.}~\cite{hui2024endow} leveraged temporal and spatial relationships between frames to generate prompts for SAM.
Different from the above approaches, we propose to handle motion cues in an explicit way as aforementioned, and a two-stream architecture is designed to conduct both optical flow estimation and segmentation.

\subsection{Video Object Segmentation (VOS) \& Video Salient Object Detection (VSOD)}

VCOD can be seen as a specialized task of video object segmentation (VOS) that segments objects across consecutive video frames.
Mei~\textit{et al.}~\cite{mei2021transvos} proposed a transformer-based framework to leverage temporal and spatial relationships across frames. To increase the detection speed, Park~\textit{et al.}~\cite{park2021learning} introduced a novel network capable of dynamically selecting mask generation methods, either by reusing features from prior frames or processing the entire network. To incorporate these advantages, memory-based networks \cite{li2022recurrent,oh2019video} have also been explored to better utilize historical information. These VOS networks use the current frame to query a memory bank storing historical features and corresponding object masks from past frames.

On the other hand, the objective of video salient object detection (VSOD) stands in direct contrast to VCOD. VSOD focuses on locating and segmenting the most prominent objects from sequences.
Chen~\textit{et al.}~\cite{chen2018scom} utilized spatial and temporal cues along with local constraints to achieve global saliency optimization.
Li~\textit{et al.}~\cite{li2018unsupervised} proposed a motion-based bilateral network for background estimation, and the background estimation results are then merged with instance embeddings into a graph, where edges connect pixels across different frames for multi-frame reasoning.
Song~\textit{et al.}~\cite{song2018pyramid} proposed a novel recurrent network to extract multi-scale spatial features, which are then concatenated and fed into an extended deep bidirectional ConvLSTM to learn spatiotemporal information.
Cong~\textit{et al.}~\cite{cong2019video} designed a two-stage method: first, obtaining the spatial saliency of each frame through sparsity-based reconstruction, and then capturing the sequential correspondence in the temporal space via progressive sparsity-based propagation.
Xu~\textit{et al.}~\cite{xu2019video} proposed a novel method for modeling motion energy based on four aspects: gradient flow field, motion direction, motion magnitude, and the spatial gradients of video frames.
Yan~\textit{et al.}~\cite{yan2019semi} ~\textit{et al.} leveraged optical flow estimation to generate pseudo-labels for some unannotated frames in the dataset, and further enhances the spatio-temporal correlation between video frames using Non-local~\cite{wang2018non} and ConvGRU~\cite{ballas2016delving}.
Chen~\textit{et al.}~\cite{chen2021exploring} integrated a lightweight temporal model into the spatial branch, coarsely locating spatial saliency regions associated with highly confident salient motion. Simultaneously, the spatial branch itself can iteratively optimize the temporal model in a multi-scale manner.
Ji~\textit{et al.}~\cite{ji2021full} proposed a full-duplex strategy to obtain more stable consistent features.
Zhao~\textit{et al.}~\cite{zhao2024motion} introduced a space-time memory-based network and leveraged high-level features to refine low-level details.

Compared to existing VOS and VSOD methods, our model excels in discerning subtle differences between camouflaged objects and their highly similar surroundings by simultaneously integrating camouflage properties with explicit motion modeling information.

\subsection{Visual Prompt Learning}
Recently, prompt learning has emerged as a new paradigm that has significantly enhanced the performance of natural language processing (NLP) tasks~\cite{brown2020language}. Besides, the prompting paradigm has been adopted in many computer vision tasks \cite{sandler2022fine,jia2022visual}. 
Work \cite{sandler2022fine} modified transformer layers by introducing memory tokens, constituting a set of learnable embedding vectors. 
VPT~\cite{jia2022visual} employed a similar strategy by applying learnable embedding vectors to transformer encoders, achieving noteworthy performance across various downstream recognition tasks. 
Based on this idea, ViPT~\cite{zhu2023visual} integrated modality-complementary visual prompts for task-oriented multi-modal tracking.
The popular Segment Anything Model (SAM) \cite{kirillov2023segment} integrated various visual prompts like points, boxes, or masks to achieve tailored object segmentation and exhibited decent zero-shot generalization.
Previous researches have mainly focused on specific tasks like classification, tracking, or segmentation. 
In this paper, inspired by \cite{zhu2023visual}, we treat intermediate features from one stream as a prompt for injecting complementary information to the other, and propose an interactive prompting scheme for the VCOD task.

\section{Methodology}\label{sec:method}
\begin{figure*}[t!]
\begin{center}
\includegraphics[width=1.0\textwidth]{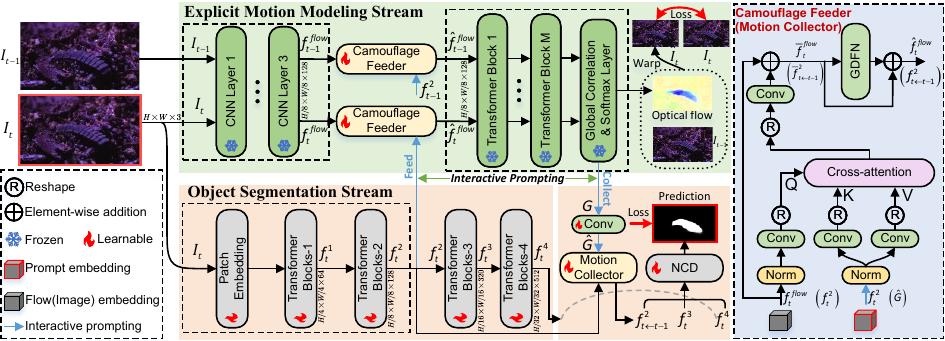}
\end{center}
\caption{\small
Overall architecture of the proposed \ourmodel, which consists of two separate streams: explicit motion modeling stream (upper) and object segmentation stream (lower). We use GMFlow~\protect\cite{xu2022gmflow} as the fundamental model to handle motion cues. With the camouflage feeder and motion collector, segmentation and motion prompts are injected into each task-specific stream to compensate essential information. 
The fire/snowflake symbols indicate that the model parameters in this part or block are designated as learnable/frozen.}
\label{fig:framework}
\end{figure*}

In this work, we propose an end-to-end trainable network EMIP to jointly optimize camouflaged object detection and optical flow estimation.
We input an adjacent frame pair $(I_t,I_{t-1})$ of a video to \ourmodel, and output a binary segmentation mask of the reference frame $I_t$, together with optical flow estimation $\boldsymbol{V}$. The overall architecture of our \ourmodel~is illustrated in Fig. \ref{fig:framework}. Note that, for object segmentation stream, a set of features $\{f_{n}^{i} \in \mathbb{R}^{H/{2^{i+1}} \times W/{2^{i+1}}\times C_i}, n\in\{t,t-1\}, i=1,...,4\}$ with different scales are extracted from a vision transformer backbone (\ie, PVTv2-B5~\cite{wang2022pvt}), the same as SLT-Net~\cite{cheng2022implicit}, for fair comparison. $W$, $H$, and $C$ represent the width, height, and channel number, respectively. Similar to~\cite{fan2020pra}, we adopt the top-three features ($f_t^{2},f_t^{3}, f_t^{4}$) for appearance modeling, and discard the first-layer feature $f_t^{1}$.

\subsection{Fundamental Model for Motion Modeling}
To achieve more effective integration of motion information for the VCOD task, we select GMFlow~\cite{xu2022gmflow} as our fundamental model, which is trained on $\sim$50k frame pairs for optical flow estimation. As shown in Fig. \ref{fig:framework}, GMFlow can be delineated into two integral components, \ie, a CNN encoder and a transformer decoder. It initially employs the CNN encoder to capture low-level features such as edges, colors, and textures. Subsequently, the transformer decoder, composed of a sequence of self- and cross-attention layers, predicts optical flow map $\boldsymbol{V}\in\mathbb{R}^{H/8\times W/8\times 2}$ and matching distribution $\boldsymbol{M}\in\mathbb{R}^{H/8\times W/8\times H/8\times W/8}$ simultaneously. 
$\boldsymbol{V}$ is derived from $\boldsymbol{M}$ through operations such as pixel-grid sampling. 
Here, we flatten the first two dimensions of $\boldsymbol{M}$ to obtain $\boldsymbol{G}\in \mathbb{R}^{H/8\times W/8\times HW/64}$, and employ $\boldsymbol{G}$ as the motion prompt for interacting with segmentation features, leveraging its detailed pixel-to-pixel matching information.
Additionally, the optical flow map $\boldsymbol{V}$ is utilized to reconstruct frame $I_t$ from $I_{t-1}$, leading to the computation of a self-supervised loss.
For more details about our fundamental motion estimation model, we refer readers to GMFlow~\cite{xu2022gmflow}. 

\subsection{Segmentation-to-Motion Prompt}
A good prompt can fully exploit the potential of the fundamental model, the same for our EMIP. Therefore, the segmentation prompt has to determine where and how to embed it in the fundamental model, considering the nature of the optical flow estimation task.

\subsubsection{Position of prompt}
The selection of the prompt position is guided by two key observations. Firstly, the choice of the fundamental model plays a pivotal role. Each fundamental model is tailored with a specific architecture; in the case of GMFlow within our \ourmodel~framework, its transformer blocks synergistically form a functional unit crafted for calculating the similarity of pixel features between two frames. Thus, preserving the integrity of this architecture becomes paramount for the coherent generation of optical flow.
Secondly, the architectural arrangement of neural networks involves a stratification where lower layers are dedicated to extracting basic features such as edges, colors, and textures, each exhibiting distinctive characteristics across various modalities \cite{zeiler2014visualizing}. This stratification has been shown by \cite{liu2020learning} to be particularly advantageous for task like optical flow estimation, emphasizing the importance of low-level visual correspondences. Leveraging this insight, we position the segmentation prompt at the lower layers of both the  segmentation and motion modeling streams, with the direction being from segmentation to motion. Further details are discussed in Section~\ref{sec: Ablation}, where we delve into various prompt position choices and their impact on results.

\subsubsection{Camouflage feeder}
To better incorporate the segmentation prompt into the motion stream, we design camouflage feeder. 
As shown in Fig. \ref{fig:framework} right, our camouflage feeder is based on cross attention with a residual connection, which takes the optical flow input feature $f^{flow}_t \in \mathbb{R}^{\frac{HW}{8^2} \times d}$ as the source of \emph{query}, and the segmentation prompt feature $f^2_{t} \in \mathbb{R}^{\frac{HW}{8^2} \times d}$ as the source of \emph{key} and \emph{value}. The residual connection is to maintain more query-related cues, and $d$ represents the embedding dimension. This process can be written as:
\begin{equation}
    {\hat{f}}^{flow}_t={\overline{f}}^{flow}_t+\mathtt{GDFN}({\overline{f}}^{flow}_t)
\end{equation}
where ${\overline{f}}^{flow}_t$ is formulated as:
\begin{equation}
    {\overline{f}}^{flow}_t=f^{flow}_t+\mathtt{CA}(\mathbf{Q},\mathbf{K},\mathbf{V}),
\end{equation}
where $\mathtt{CA}$ is defined as:
\begin{equation}
    \mathtt{CA}(\mathbf{Q},\mathbf{K},\mathbf{V})=\mathtt{Softmax}(\mathbf{Q}\mathbf{K}^\intercal/\sqrt{d})\mathbf{V},
\end{equation}
and $\mathbf{Q},\mathbf{K},\mathbf{V}\in \mathbb{R}^{\frac{HW}{8^2} \times d}$ represent \emph{query}, \emph{key}, and \emph{value} matrices, respectively. 
$\bf Q/\bf K$ and $\bf V$ are derived by applying layer normalization and 3$\times$3 convolution to the input  $f^{flow}_t$ and $f^2_t$, respectively.
Here, we employ a Gated-Dconv Feed-forward Network (GDFN)~\cite{zamir2022restormer} to suppress the aggregation of noisy prompt, and GDFN for input $\mathbf{X}$ is defined as:
\begin{equation}
    \mathtt{GDFN}(\mathbf{X})=\mathtt{C1}\left[\phi(\mathtt{C3}(\mathtt{C1}(\mathtt{LN}(\mathbf{X}))))\odot \mathtt{C3}(\mathtt{C1}(\mathtt{LN}(\mathbf{X})))\right],
\end{equation}
where $\mathtt{C1}$ and $\mathtt{C3}$ are 1$\times$1 convolution and 3$\times$3 depth-wise convolution, respectively. $\phi$ is GELU function~\cite{hendrycks2016gaussian} and $\odot$ denotes element-wise multiplication. 
Note that, $\mathtt{GDFN}$ comprises two separate paths, one of which is activated with GELU function. Subsequently, an element-wise product is applied between these two paths. 
Literature~\cite{zamir2022restormer} has demonstrated that the gating mechanism in GDFN can control information flow and yield better performance compared with the conventional feed-forward network (FN)~\cite{vaswani2017attention} for feature restoration. For the same purpose, to mitigate the impact of the integrated noisy prompts, we employ GDFN to control noisy information flow.
Finally, the resulting segmentation-prompted motion features $\hat{f}^{flow}_t$ with both knowledge of motion and appearance can be well adapted to subsequent motion reconstruction. 

\subsubsection{Prompt learning}
To better learn the motion information, inspired by \cite{liu2021amd}, we employ a self-supervised loss to optimize the motion-to-segmentation prompt learning. Further elaboration is provided in Section~\ref{sec: Loss}.

\subsection{Motion-to-Segmentation Prompt}
Motion-to-segmentation prompt formally provides an auxiliary motion flow that is both temporally and spatially coordinated with the segmentation stream.
After obtaining the segmentation-prompted motion, we need to collect this knowledge back to the segmentation stream. Thus, we introduce another module named Motion Collector to integrate motion prompts into the segmentation stream. Given the functional analogy of motion collector with the camouflage feeder, we opt to maintain the structural configuration of the interaction block rather than introducing a redesign (as illustrated in Fig.~\ref{fig:framework} right). Specifically, upon receiving motion feature $\boldsymbol{G}$ (flattening the first two dimensions of $\boldsymbol{M}$), we then apply a 3$\times$3 convolution operation to $\boldsymbol{G}$ to adjust and reduce its channel dimension to align with $f^{2}_t$. 
Then we feed the motion-to-segmentation prompt $\hat{\boldsymbol{G}}=\mathtt{Conv}(\boldsymbol{G})$ and $f^{2}_t$ to the motion collector to obtain the prompted appearance feature $f_{t \leftarrow t-1}^{2}$.

Subsequently, following that most VCOD works \cite{ji2023gradient,cheng2022implicit} use neighbor connection decoder (NCD)~\cite{cheng2022implicit} to predict the segmentation map, we use NCD to decode the appearance features. NCD with motion collector can be regarded as the prediction head for the motion fundamental model, similar to the concepts of some previous prompt learning methods~\cite{ji2023gradient,cheng2022implicit} that design an extra prediction head after freezing the backbone parameters. 
Specifically, the appearance feature $f_{t \leftarrow t-1}^{2}$ prompted by motion combined with $f^3_t$ and $f^4_t$ are fed to NCD to obtain the predicted map. 
The optimization of motion-to-segmentation prompt and prediction head is under the constraint of segmentation loss, and more details will be given in Sec.~\ref{sec: Loss}.

\textbf{Discussion.}
The motivation for our prompt learning strategy, freezing the motion stream and fully tuning the segmentation stream, is from the observed performance degradation of the PVT~\cite{wang2022pvt} pre-trained on ImageNet when directly applied to the VCOD task. Similar challenges are also encountered with large models like SAM~\cite{kirillov2023segment}, which struggle in camouflaged scenes~\cite{ji2023sam}. Currently, there lacks a freezable foundational model that excels on the VCOD task. In contrast, we find that the optical flow model could achieve good generalization in most VCOD scenarios, when most of its parameters are frozen.

\subsection{Supervision and Loss Function}
\label{sec: Loss}
To ensure optimization for each component of the model, we separately define loss functions for the motion and segmentation stream, and then use these two losses  to optimize the entire model jointly.
Considering the lack of ground truth (GT) optical flow in video camouflaged scenarios, we propose a self-supervised strategy to learn the optical flow estimation stream. For the sake of brevity and clarity, we refer to \cite{liu2021amd} and warp the frame $I_{t-1}$ according to the optical flow estimation $\boldsymbol{V}$ to obtain the reconstructed reference frame for $I_t$, which is denoted as $\hat{I}_t$, \ie, $\hat{I}_t=\mathtt{Warp}(\boldsymbol{V},I_{t-1})$. We supervise the optical flow by computing the distance between $I_t$ and $\hat{I}_t$. Thus, the flow loss $\mathcal{L}_{\text{flow}}$ can be expressed as:
 \begin{equation}
    \mathcal{L}_{\text{flow}} = \mathtt{SSIM}(I_t,\hat{I}_t),
\end{equation}
where $\mathtt{SSIM}$ is pixel-wise photometric loss~\cite{wang2004image}, which is commonly used for reflecting image distortion from three aspects: brightness, contrast, and structure.

For the segmentation stream, we employ a hybrid loss function \cite{21Fan_HybridLoss}, the same as SLT-Net~\cite{cheng2022implicit}. The hybrid loss $\mathcal{L}_{\text{seg}}$ is defined as $\mathcal{L}_{\text{seg}} = \mathcal{L}_{\text{IoU}} + \mathcal{L}_{\text{bce}} + \mathcal{L}_{\text{e-loss}}$, which includes IoU loss, binary cross-entropy loss, and enhanced-alignment loss.
Finally, we jointly optimize the motion modeling and object segmentation streams, and the total loss is formulated as:

\begin{equation}
\mathcal{L}_{\text{total}}=\mathcal{L}_{\text{seg}} + \mathcal{L}_{\text{flow}}.
\end{equation}

\subsection{Long-term Consistency Modeling}
\label{sec: long-term}

\begin{figure*}[t!]
\begin{center}
\includegraphics[width=1.0\textwidth]{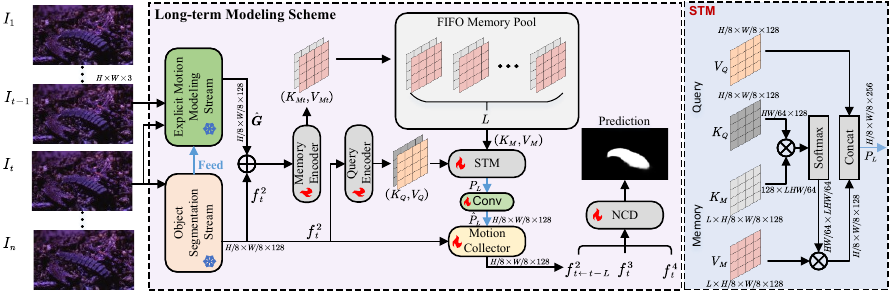}
\end{center}
\caption{Overview of our long-term modeling scheme (EMIP$^\dag$). EMIP$^\dag$ consists of a frozen EMIP and other five learning modules (\ie, Memory Encoder, Query Encoder, STM, Motion Collector, and NCD).}
\label{fig:long_term}
\end{figure*}

The short-term model captures correlated information between adjacent frames, but due to the impact of noises and the limitation of motion estimation model, the short-term model may be less robust. 
To address this, long-term consideration can be further introduced to improve results' consistency and therefore detection accuracy.
To utilize long-term historical information, as well as maintain feasible computational cost, a dynamic memory model is designed to extract a long-term prompt, which formulates historical features into the prompt to mitigate short-term prediction errors. 

Fig. \ref{fig:long_term} illustrates our long-term consistency modeling scheme, termed EMIP$^\dag$.
Specifically, during training, we freeze the short-term model (referring to the explicit motion modeling stream and the object segmentation stream in Fig. \ref{fig:long_term}) after it converges and then add the long-term memory module for further learning. 
Given a video sequence \{${{I}_{1},...,{I}_{t-1},{I}_{t}}, t>1$\}, we sequentially input current frame ${I}_{t}$ and preceding frame ${I}_{t-1}$ into the short-term model to obtain the appearance features $f^2_{t}$ and the motion-to-segmentation prompt $\hat{\boldsymbol{G}}$ between the two frames. 
The element-wise sum of $f^2_{t}$ and $\hat{\boldsymbol{G}}$ is conducted, and the results are then fed to the memory encoder to be mapped as key $K_{Mt}$ and value $V_{Mt}$. 
The memory encoder consists of a convolutional layer, followed by a normalization layer and a ReLU activation function, and then two parallel convolutional layers. 
The input features are processed to obtain the key-value mappings $K_{Mt}$ and $V_{Mt}$. The processed key-value mappings are later stored into the memory pool.

Considering the computational burden and minor impact of distant frames, we implement a FIFO (First-In First-Out) memory pool with fixed capacity ${L}$. 
Note that such a fixed capacity is configurable according to practical applications.
Referring to the long-term setting of SLT-Net~\cite{cheng2022implicit}, we set ${L}$ to 5 in this paper.
The object feature $f^2_{t}$ extracted from the current frame are fed to the query encoder, which consists of two parallel convolutional layers used to compute the query key-value mappings $K_Q$ and $V_Q$. 
These mappings are then used to query the memory pool, which stores appearance and motion features for both the current and historical frames, via a space-time memory read block (STM)~\cite{oh2019video} (referring to the right part of Fig.~\ref{fig:long_term}). This process yields the long-term prompt ${P}_{L}$. 
Specifically, the key-value mappings of each historical frame in the memory pool are concatenated along the temporal dimension to form $K_M$ and $V_M$. These concatenated mappings are then interacted with the query key-value mappings $K_Q$ and $V_Q$ through the STM module, ultimately producing the query results.
Next we apply a $3\times 3$ convolution operation to ${P}_{L}$ to adjust and reduce its channel dimension to align with $f^2_{t}$.
Then, $\hat{P}_{L}=\mathtt{Conv}({P}_{L})$ and $f^2_{t}$ are fed to the motion collector\footnote{Can re-use the same motion collector of the short-term period if the short-term output is no longer needed.} for interaction, and the output of collector, denoted as $f_{t \leftarrow t-L}$, is subsequently fed to NCD for decoding. 

We still adopt the hybrid loss $\mathcal{L}_{\text{seg}}$, the same as the short-term version, for training the long-term counterpart.
It is worth noting that compared with SLT-Net which defines long-term modeling as a sequence-to-sequence problem (requiring all frames of a clip), our scheme utilizes only historical frames regarding the current frame. Hence, our scheme is theoretically more suitable for real-time and practical applications, where only past information is available.

\section{Experiments and Results}\label{sec:exp}
\subsection{Datasets and Metrics}
\subsubsection{Datasets}
Following \cite{cheng2022implicit,hui2024implict-explicit}, we conduct experiments on two widely recognized VCOD benchmarks: MoCA-Mask~\cite{cheng2022implicit} and CAD~\cite{bideau2016s}. Among them, MoCA-Mask stands as the most challenging dataset, comprising 19,313 frames across 71 clips for training, and 3,626 frames of 16 clips for testing. The CAD dataset includes 836 frames of 9 clips designated for testing.
To assess the generalizability of our model, we conducted experiments on four widely used VSOD/VOS datasets: DAVIS$_{16}$~\cite{perazzi2016benchmark} (with 30 training clips and 20 testing clips), FBMS~\cite{ochs2013segmentation} (29 training clips and 30 testing clips), ViSal~\cite{wang2015consistent} (comprising 17 video sequences for testing) and SegV2~\cite{li2013video} (including 13 clips for testing).

\subsubsection{Evaluation Metrics}
We adopt widely recognized evaluation metrics to assess our model performance, namely: structure measure ($\mathcal{S}_\alpha$)~\cite{fan2017structure}, weighted F-measure ($F_{\beta}^{w}$)~\cite{margolin2014evaluate}, F-measure (max $\mathcal{F}$)~\cite{achanta2009frequency}, mean absolute error ($\mathcal{M}$)~\cite{perazzi2012saliency}, and mean value of Dice and IoU. These metrics provide a comprehensive and reliable assessment of model performance.

\subsection{Implementation Details}
For fair comparisons, we adhere to the training settings outlined in \cite{cheng2022implicit} and employ PVTv2~\cite{wang2022pvt} as the feature extraction backbone. All input images are resized to 352$\times$352 and subject to data augmentation techniques, including color enhancement and random rotation.
A two-stage training pipeline in \cite{cheng2022implicit} is also adopted by our model: first train the backbone on the static training set of COD10K (3,040 images)~\cite{fan2021concealed}, and then fine-tune the whole model with the temporal components on the training set of MoCA-Mask (19,313 frames)~\cite{cheng2022implicit}. 
The entire model is optimized using the Adam optimizer~\cite{kingma2014adam} with a cosine annealing strategy. The maximum and minimum learning rates, along with the maximum adjusted iterations, are set to 1e-5, 1e-6, and 20, respectively.
EMIP is trained for 7 hours over 60 epochs on an NVIDIA 4090 GPU with 16,536 MB of memory using a batch size of 6.
For inference, \ourmodel~takes any two consecutive frames as input and outputs the segmentation prediction together with the corresponding optical flow estimation. However, for EMIP$^\dag$, it requires consecutive-frame pairs to be input sequentially, in order to leverage its long-term property.
For VSOD, we train our model on the training set of DAVIS$_{16}$ (30 clips) and FBMS (29 clips) as~\cite{ji2021full}. The proposed \ourmodel~is implemented by PyTorch~\cite{paszke2019pytorch}, and all experiments are conducted on an NVIDIA RTX 4090 GPU.

\begin{table*}[ht!]
  \caption{Quantitative comparisons with state-of-the-art methods on MoCA-Mask and CAD datasets. $\dag$ denotes the long-term version. ``$\uparrow$" / ``$\downarrow$" indicates that larger/smaller is better. Top three results are highlighted in \textcolor{red}{red}, \textcolor{blue}{blue} and \textcolor{indiagreen}{green}.} 
  \footnotesize
  \centering
  \tabcolsep=0.20cm
  \renewcommand{\arraystretch}{1.1}
  \begin{tabular}{r | r | c | c || ccccc | ccccc } 
  \toprule
  & & & & \multicolumn{5}{c|}{MoCA-Mask}  & \multicolumn{5}{c}{CAD}\\
  \cline{3-14}
  Method & Model & Publication & Backbone & $\mathcal{S}_\alpha\uparrow$ &$F_\beta^w\uparrow$ & $\mathcal{M}\downarrow$ & Dice$\uparrow$ & IoU$\uparrow$
  & $\mathcal{S}_\alpha\uparrow$ &$F_\beta^w\uparrow$ & $\mathcal{M}\downarrow$ & Dice$\uparrow$ & IoU$\uparrow$\\
  \hline
  \multirow{14}{*}{\emph{Image-based}} 
  &EGNet~\cite{zhao2019EGNet}& ICCV'19 & VGG-16 & 0.547 & 0.110 & 0.035 & 0.143 & 0.096 & 0.619 & 0.298 & 0.044  & 0.324 & 0.243\\
  &BASNet~\cite{Qin_2019_CVPR} & CVPR'19 & ResNet-34 & 0.561 & 0.154 & 0.042 & 0.190 & 0.137 & 0.639 & 0.349 & 0.054  & 0.393 & 0.293\\
  &CPD~\cite{Wu_2019_CVPR} & CVPR'19 & VGG-16 & 0.561 & 0.121 & 0.041 & 0.162 & 0.113 & 0.622 & 0.289 & 0.049  & 0.330 & 0.239\\
  &PraNet~\cite{fan2020pra} & MICCAI'20 & Res2Net-50 & 0.614 & 0.266 & 0.030 & 0.311 & 0.234 & 0.629 & 0.352 & 0.042  & 0.378 & 0.290\\
  &SINet~\cite{fan2020Camouflage} & CVPR'20 & ResNet-50 & 0.598 & 0.231 & 0.028 & 0.276 & 0.202 &  0.636 & 0.346 & 0.041 & 0.381 & 0.283\\
  &SINet-V2~\cite{fan2021concealed} & TPAMI'22 & ResNet-50 & 0.588 & 0.204 & 0.031 & 0.245 & 0.180 &  0.653 & 0.382 & 0.039 & 0.413 & 0.318 \\
  &ZoomNet~\cite{pang2022zoom} & CVPR'22 & ResNet-50 & 0.582 & 0.211 & 0.033 & 0.224 & 0.167 &  0.587 & 0.225 & 0.063 & 0.246 & 0.166 \\
  &DGNet~\cite{ji2023gradient} & MIR'23 & PVT & 0.581 & 0.184 & 0.024 & 0.222 & 0.156 & 0.686 & 0.416 & 0.037 & 0.456 & 0.340 \\
  &FEDER~\cite{he2023FEDER} & CVPR'23 & ResNet-50 & 0.560 & 0.165 & 0.031 & 0.194 & 0.137 & 0.691 & 0.444 & \textcolor{indiagreen}{0.029} & 0.474 & 0.375 \\
  &FSPNet~\cite{huang2023feature} & CVPR'23 & ViT & 0.594 & 0.182 & 0.044 & 0.238 & 0.167 & 0.539 & 0.220 & 0.145 & 0.309 & 0.212 \\
  &PUENet~\cite{zhang2023predictive} & TIP'23 & ViT & 0.594 & 0.204 & 0.037 & 0.302 & 0.212 & 0.673 & 0.427 & 0.034 & \textcolor{indiagreen}{0.499} & 0.389 \\
  &HitNet~\cite{hu2023high} & AAAI'23 & PVT & 0.623 & 0.299 & \textcolor{indiagreen}{0.019} & 0.318 & 0.254 & 0.685 & 0.463 & 0.031 & 0.478 & 0.373 \\
  &FSEL~\cite{sun2024frequency} & ECCV'24 & PVT & 0.596 & 0.260 & 0.053 & 0.219 & 0.151 & 0.649 & 0.368 & 0.053 & 0.434 & 0.325 \\ 
  &HGINet~\cite{Yao2024HGINet} & TIP'24 & ViT & 0.610 & 0.251 & 0.030 & 0.303 & 0.221 & 0.680 & 0.437 & 0.050 & 0.501 & 0.392 \\ 
  
  \hline
  \multirow{8}{*}{\emph{Video-based}} &
  {RCRNet} \cite{yan2019semi} & ICCV'19 & ResNet-50 & 0.555 &  0.138 &  0.033 &  0.171 & 0.116 & 0.627 & 0.287 & 0.048 & 0.309 & 0.229\\
  &{MG} \cite{yang2021selfsupervised} & ICCV'21 & VGG-style &  0.530 & 0.168 &  0.067  & 0.181  & 0.127 & 0.594 & 0.336 & 0.059 & 0.368 & 0.268\\
  &{PNS-Net} \cite{ji2021progressively} & MICCAI'21 & Res2Net-50 & 0.544 & 0.097 & 0.033 & 0.121 & 0.101 & 0.655 & 0.325 & 0.048 &  0.384 &  0.290\\
  &SLT-Net \cite{cheng2022implicit} & CVPR'22 & PVT & 0.637 & 0.304 & 0.027 & 0.356 & 0.271 & 0.696 & 0.471 & 0.031 & 0.484 & 0.392  \\
  &SLT-Net$^\dag$ \cite{cheng2022implicit} & CVPR'22 & PVT & 0.631 & 0.311 & 0.027 & 0.360 & 0.272 & \textcolor{indiagreen}{0.696} & \textcolor{indiagreen}{0.481} & 0.030 & 0.493 & \textcolor{indiagreen}{0.401}\\
  &IMEX~\cite{hui2024implict-explicit} & TMM'24 & ResNet-50 & \textcolor{indiagreen}{0.661} & \textcolor{indiagreen}{0.371} & 0.020 & \textcolor{indiagreen}{0.409} & \textcolor{indiagreen}{0.319} & 0.684 & 0.452 & 0.033 & 0.469 & 0.370  \\
  \rowcolor{ours}
  &\textbf{\ourmodel~\textit{(Ours)}}
  & -- & PVT & \textcolor{blue}{0.669} & \textcolor{blue}{0.374} & \textcolor{blue}{0.017} & \textcolor{blue}{0.424} & \textcolor{blue}{0.326} & \textcolor{blue}{0.710} & \textcolor{blue}{0.504} & \textcolor{blue}{0.029} & \textcolor{blue}{0.528} & \textcolor{blue}{0.415}\\
  \rowcolor{ours}
  &\textbf{\ourmodel$^\dag$~\textit{(Ours)}}
  & -- & PVT & \textcolor{red}{0.675} & \textcolor{red}{0.381} & \textcolor{red}{0.015} & \textcolor{red}{0.426} & \textcolor{red}{0.333} & \textcolor{red}{0.719} & \textcolor{red}{0.514} & \textcolor{red}{0.028} & \textcolor{red}{0.536} & \textcolor{red}{0.425}\\
  
  \bottomrule
  \end{tabular}    
  \label{tab:Moca}
\end{table*}

\begin{figure*}[t!]
    \centering
    \includegraphics[width=1.0\textwidth]{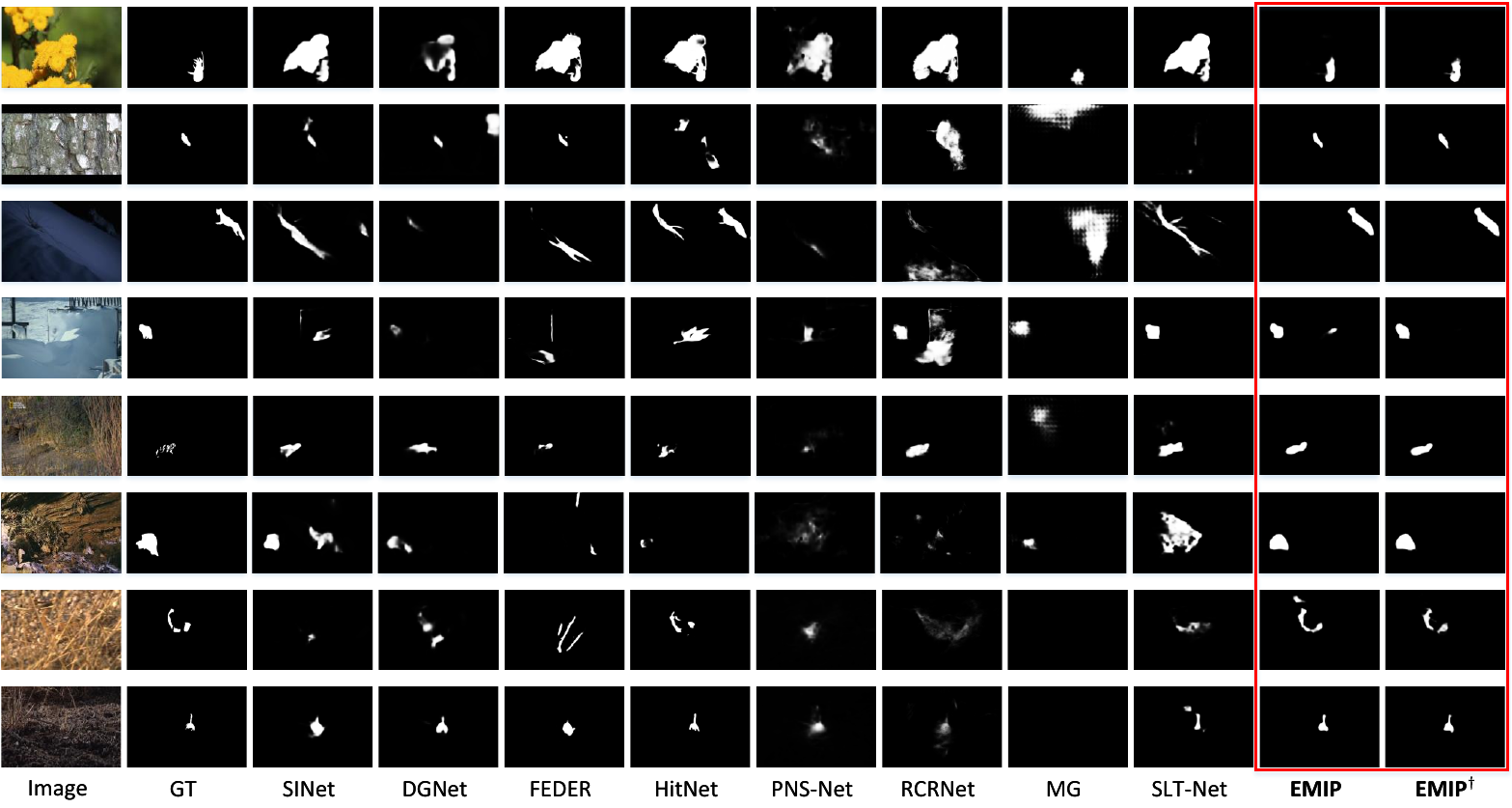}
    \caption{Visual comparisons of our models (EMIP and EMIP$^\dag$) with eight state-of-the-art methods. We select some difficult scenarios, including dusky night, fast-moving objects, stationary objects, small objects, and noisy backgrounds.}
    \label{fig: visualization_result}
\end{figure*}

\subsection{Comparisons with State-of-the-arts}
\label{sec:Comparison with State-of-the-Arts}
To demonstrate the effectiveness of our \ourmodel, we compare it with various state-of-the-arts. These compared methods can be categorized into two types: (i) Image-based camouflaged object detection methods \cite{zhao2019EGNet,Qin_2019_CVPR,Wu_2019_CVPR,fan2020pra,fan2020Camouflage,fan2021concealed,pang2022zoom,ji2023gradient,he2023FEDER,huang2023feature,zhang2023predictive,hu2023high}, which are designed for detecting objects in static camouflaged scenes; (ii) Video-based object detection methods \cite{yan2019semi,yang2021selfsupervised,ji2021progressively,cheng2022implicit,hui2024implict-explicit}, which focus on identifying camouflaged or moving objects in dynamic video sequences. Notably, we include comparisons with SLT-Net~\cite{cheng2022implicit} and IMEX~\cite{hui2024implict-explicit}, two most relevant state-of-the-art approaches for video camouflaged object detection. For fair evaluation, the predictions from these methods are either directly downloaded from the official repositories or generated using open-source code provided by the original authors.

\subsubsection{Quantitative Evaluation}
Table~\ref{tab:Moca} presents the experimental results on VCOD datasets. Notably, on MoCA-Mask dataset, our method demonstrates significant improvements:
(i) It surpasses the previous state-of-the-art image-based approach, HitNet~\cite{hu2023high}, by 27\% in $F_\beta^w$, underscoring the efficacy of incorporating motion cues into the VCOD task.
(ii) Additionally, when compared to SLT-Net~\cite{cheng2022implicit}, which utilizes the same backbone as ours, our method shows a marked improvement, achieving a 20\% gain in $F_\beta^w$. Although our model was trained on the MoCA-Mask dataset, the testing results on the CAD dataset underscore its superior robustness and quantitative performance enhancements. The notable performance gains of \ourmodel~over existing and recently proposed techniques emphasize that the combination of explicit motion modeling and prompt learning substantially enhances the completeness of detected camouflaged objects.

Additionally, by leveraging long-term temporal information, our model achieves state-of-the-art performance across all five evaluation metrics on both VCOD benchmarks. This indicates that preserving long-term consistency effectively suppresses motion noise of camouflaged objects and enhances the stability of video predictions. Compared to SLT-Net$^\dag$~\cite{cheng2022implicit} and IMEX~\cite{hui2024implict-explicit}, which also utilizes long-term consistency information as external cues to refine predictions, our long-term strategy yields notable performance gains.
It is worth noting that, unlike SLT-Net, which frames long-term modeling as a sequence-to-sequence problem (requiring all frames of a clip), our approach utilizes only historical frames relative to the current frame. This design makes our scheme theoretically more suitable for real-time and practical applications, where only past information is accessible.

\subsubsection{Qualitative Comparison}
\begin{figure}[t!]
\begin{center}
\includegraphics[width=.481\textwidth]{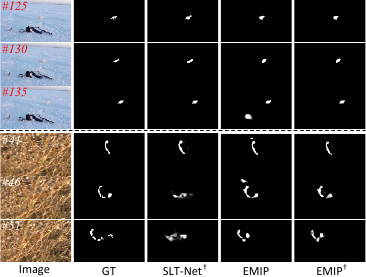}
\end{center}
\vspace{-8pt}
\caption{Visual comparisons on consecutive video frames.}
\label{fig:visual_long_term}
\end{figure}

Fig. \ref{fig: visualization_result} illustrates qualitative comparisons by visualizing the segmentation results of several examples. Our model demonstrates superior alignment with ground truth, showcasing its enhanced capability in identifying camouflaged objects compared to other methods. 
Furthermore, the segmentation results over consecutive frames of the same clip are shown in Fig. \ref{fig:visual_long_term}. One can see that incorporating the long-term modeling scheme, namely EMIP$^\dag$, can reduce errors in short-term prediction and lead to boosted performance.
Fig. \ref{fig:Visualization_of} presents visual comparisons of optical flow prediction in camouflaged scenarios. The designed prompting strategy in our model ensures more precise responses within camouflaged regions, while minimizing the influence of irrelevant motion noise. 
Consequently, our model excels in concurrently performing the motion modeling and segmentation tasks, leveraging the integration of enhanced features to improve the detection of camouflaged objects.

\subsection{Ablation Studies}\label{sec: Ablation}
To evaluate the effectiveness of each core component, we conduct thorough ablation studies by removing or substituting components from the complete EMIP.

\subsubsection{Camouflage feeder and motion collector}
\begin{figure}[h!]
\begin{center}
\includegraphics[width=.481\textwidth]{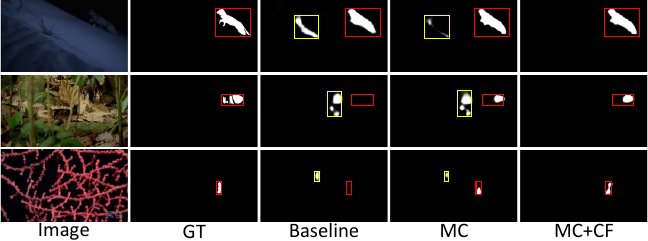}
\end{center}
\vspace{-8pt}
\caption{Qualitative results of CF and MC. The red box indicates the ground-truth location of the camouflaged object, while the yellow box represents detected noise. With the incorporation of MC, the camouflaged object can be effectively identified. Moreover, integrating CF further suppresses noise.}
\vspace{-8pt}
\label{fig:Camouflage_feeder_and_motion_collector}
\end{figure}

As shown in Table~\ref{tab:ablation of each module}, we first validate the role of the two prompt integration modules, \ie, camouflage feeder and motion collector. 
We start with the baseline model (\#1), which directly uses the segmentation stream to extract single-frame information for prediction, \ie, removing the motion modeling stream of EMIP. 
Configuration \#2 means using only motion information to prompt the segmentation stream, meanwhile discarding segmentation-to-motion prompt. 
Model \#3 is the full EMIP.
Comparing configuration \#2 with \#1, it is evident that camouflaged objects are challenging to detect without inter-frame motion information. In scenes with movement, motion information provides additional context that can improve the accuracy of segmentation. It helps in identifying and separating camouflaged objects that are in motion, which might be challenging using appearance-based methods alone. 
Comparing configuration \#3 with \#2, we observe that incorporating the segmentation-to-motion prompt via the camouflage feeder into the motion stream enhances the robustness of motion cues. This improvement in motion cue robustness subsequently leads to superior performance in camouflaged object detection.
The visualization results are presented in Fig.~\ref{fig:Camouflage_feeder_and_motion_collector}. 
Relying solely on appearance features makes it challenging to accurately detect or localize truly camouflaged objects in certain scenarios.
With the introduction of the motion collector, the camouflaged object is detected using motion information. Furthermore, incorporating the camouflage feeder to inject camouflage priors effectively reduces interference and enables more robust localization of the camouflaged object.

\begin{table}[h!]
\caption{Ablation results on camouflage feeder~(CF) and motion collector~(MC) modules of \ourmodel~on MoCA-Mask dataset. The best results are highlighted in \textbf{bold}.}
\centering
\footnotesize
\renewcommand{\arraystretch}{1.1}
\renewcommand{\tabcolsep}{0.23cm}
\begin{tabular}{c|cc||ccccc}
    \toprule
    \# & CF & MC &
    $\mathcal{S}_\alpha\uparrow$ &$F_\beta^w\uparrow$ & $\mathcal{M}\downarrow$ & Dice$\uparrow$ & IoU$\uparrow$ \\
    \hline
    1 & - & - & 0.627 & 0.268 & 0.031 & 0.340 & 0.254 \\
    2 & - & \Checkmark &0.657 & 0.349 & 0.020 & 0.394 & 0.300  \\
     3 & \Checkmark & \Checkmark
     & \textbf{0.669} & \textbf{0.374} & \textbf{0.017} & \textbf{0.424} & \textbf{0.326}\\
    \bottomrule
\end{tabular}
\vspace{-4pt}
\label{tab:ablation of each module}
\end{table}

\subsubsection{Full-tuning \emph{v.s.} Freezing}
\begin{figure}[t!]
\begin{center}
\includegraphics[width=.481\textwidth]{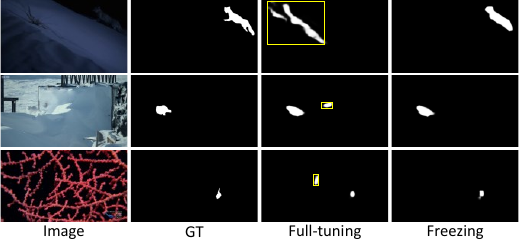}
\end{center}
\vspace{-7pt}
\caption{Qualitative results of full-tuning and freezing the motion stream. The yellow boxes represent false positive predictions.}
\label{fig:Full-tuning_vs_Freezing}
\end{figure}

Existing models usually address motion modeling by fully tuning all the parameters. In contrast, our EMIP adopts a prompt learning strategy to conduct motion modeling by freezing the motion fundamental model. From the comparisons in Table~\ref{tab: frozen}, where both settings are preloaded with pre-trained weights, the superiority of freezing the motion model over full-tuning is evident. 
As can be seen from Fig.~\ref{fig:Full-tuning_vs_Freezing}, full-tuning the motion stream disrupts its ability to effectively model motion, introducing additional noise into camouflaged object prediction.
Freezing all motion modeling layers allows the model to leverage robust and general features learned from large-scale motion estimation datasets during pre-training. This strategy ensures that the prompting process concentrates on learning camouflage-specific features, thereby enhancing the efficiency and effectiveness of the optimization process of our \ourmodel. As demonstrated in Table~\ref{tab: frozen}, the prompt learning strategy within our framework for the motion modeling stream significantly exploits its potential on limited VCOD data, even in the absence of ground truth for optical flow.

\begin{table}[t!]
    \caption{Quantitative comparison of full-tuning and freezing the explicit motion modeling stream.}
    \footnotesize
    \centering
    \tabcolsep=0.23cm
    \renewcommand{\arraystretch}{1.1}
    \begin{tabular}{c||ccccc}
    \toprule
      Setting & $\mathcal{S}_\alpha\uparrow$ &$F_\beta^w\uparrow$ & $\mathcal{M}\downarrow$ & Dice$\uparrow$ & IoU$\uparrow$ \\
     \hline
     Full-tuning & 0.645 & 0.326 & 0.018 & 0.364 & 0.281\\
     Freezing \textit{(Ours)} & \textbf{0.669} & \textbf{0.374} & \textbf{0.017} & \textbf{0.424} & \textbf{0.326} \\
    \bottomrule
    \end{tabular}
    \vspace{-7pt}
    \label{tab: frozen}
\end{table}

\subsubsection{Prompt destination on the motion stream}
\begin{figure}[h!]
\begin{center}
\includegraphics[width=.481\textwidth]{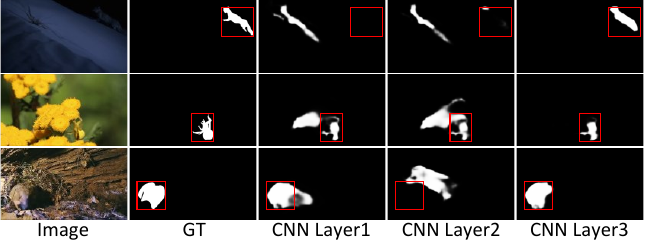}
\end{center}
\vspace{-8pt}
\caption{Visualization of different prompt destinations on the motion stream. The red box indicates the ground-truth location of the camouflaged object.}
\label{fig:Prompt_destination_on_the_motion_stream}
\end{figure}

The prompt destination is crucial as it determines where to introduce the appearance prompt within the motion fundamental model. In this experiment, we selected $f^2_t$ to serve dual roles: as the segmentation-to-motion prompt and as the input for the motion collector. Subsequently, we tested its effectiveness by prompting various convolutional neural network (CNN) layers within the model. The specific layers chosen for this evaluation ranged from initial to deeper CNN layers, allowing us to observe the impact at different stages. The evaluation results shown in Table~\ref{tab: prompt position} reveal that prompting with the features immediately after CNN Layer3 yields the best performance among all tested positions. 
Fig.~\ref{fig:Prompt_destination_on_the_motion_stream} shows that injecting camouflage priors as prompts after CNN Layer3, but before motion relationship modeling, effectively mitigates the interference of inaccurate motion noise.
This observation suggests that the features extracted by CNN Layer3 strike an optimal balance between low-level details and high-level abstractions. The superior performance can be attributed to the necessity for distinct-level features to be complemented by corresponding prompts that enhance level-specific representations. Essentially, the matching prompt at this stage helps in refining and emphasizing the critical features pertinent to motion and appearance, thereby boosting the overall performance of the model.
By ensuring that the prompt is aligned with the level-specific characteristics of the features, our model can more effectively capture and utilize the nuanced information present at this stage. This experiment illustrates the importance of precise prompt positioning in the context of motion and appearance integration.

\begin{table}[h!]
    \caption{Quantitative comparison of different prompt destinations on the motion stream.}
    \footnotesize
    \centering
    \tabcolsep=0.23cm
    \renewcommand{\arraystretch}{1.1}
    \begin{tabular}{c||ccccc}
    \toprule
      Prompt position & $\mathcal{S}_\alpha\uparrow$ &$F_\beta^w\uparrow$ & $\mathcal{M}\downarrow$ & Dice$\uparrow$ & IoU$\uparrow$ \\
     \hline
     CNN Layer1 & 0.644 & 0.331 & 0.018 & 0.368 & 0.285\\
     CNN Layer2 & 0.650 & 0.334 & 0.021 & 0.379 & 0.289\\
     CNN Layer3 & \textbf{0.669} & \textbf{0.374} & \textbf{0.017} & \textbf{0.424} & \textbf{0.326} \\
    \bottomrule
    \end{tabular}
    \vspace{-14pt}
    \label{tab: prompt position}
\end{table}

\subsubsection{Prompt source from the segmentation stream}
\begin{figure}[h!]
\begin{center}
\includegraphics[width=.481\textwidth]{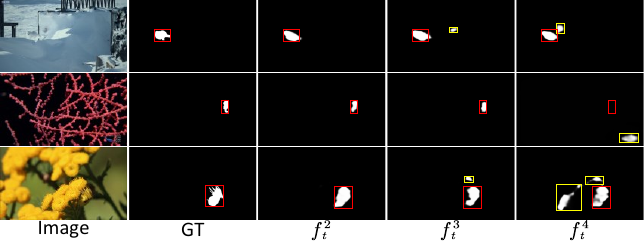}
\end{center}
\vspace{-8pt}
\caption{Qualitative results of different prompt sources from the segmentation stream. The red box denotes the ground-truth location of the camouflaged object, while the yellow box highlights the detected noise.}
\vspace{-8pt}
\label{fig:Prompt_source_from_the_segmentation_stream}
\end{figure}

To thoroughly investigate the most suitable appearance prompt to feed to the motion stream, we conducted an additional set of experiments using various prompt features, \ie, $f^2_t$, $f^3_t$, and $f^4_t$. The experimental results are summarized in Table~\ref{tab: posion of camouflage feeder} and 
Fig.~\ref{fig:Prompt_source_from_the_segmentation_stream}
, clearly demonstrating that employing appearance features $f^2_t$ as the prompt leads to superior performance. 
A plausible explanation for this observation is that subtle movements of objects in motion are likely to be overlooked in lower-resolution feature maps, such as $f^3_t$ and $f^4_t$. Consequently, this omission diminishes the efficacy of these features when used as prompts in the motion stream, thereby underscoring the importance of higher-resolution appearance features in capturing fine-grained motion details.

\begin{table}[h!]
    \caption{Different prompt sources from the segmentation stream.}
    \footnotesize
    \centering
    \tabcolsep=0.29cm
    \renewcommand{\arraystretch}{1.1}
    \begin{tabular}{c||ccccc}
    \toprule
      Features & $\mathcal{S}_\alpha\uparrow$ &$F_\beta^w\uparrow$ & $\mathcal{M}\downarrow$ & Dice$\uparrow$ & IoU$\uparrow$ \\
     \hline
     $f^2_t$ & \textbf{0.669} & \textbf{0.374} & \textbf{0.017} & \textbf{0.424} & \textbf{0.326} \\
     $f^3_t$& 0.654 & 0.350 & 0.018 & 0.389 & 0.300 \\
     $f^4_t$ & 0.624 & 0.289 & 0.023 & 0.331 & 0.251 \\
    \bottomrule
    \end{tabular}
    \vspace{-6pt}
    \label{tab: posion of camouflage feeder}
\end{table}

\subsubsection{Prompt destination on the segmentation stream}
\begin{figure}[h!]
\vspace{-16pt}
\begin{center}
\includegraphics[width=.481\textwidth]{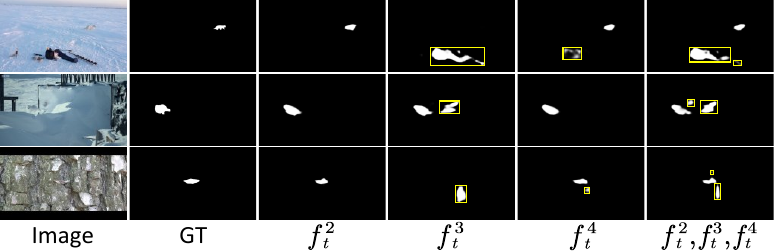}
\end{center}
\vspace{-8pt}
\caption{Qualitative results of different prompt destinations on the segmentation stream. 
The yellow boxes represent false positive detections.}
\label{fig:Prompt_destination_on_the_segmentation_stream}
\end{figure}

We harness motion information to selectively prompt distinct appearance features or to simultaneously prompt all features by incorporating additional motion collectors into the model. The empirical results, as presented in Table~\ref{tab: posion of motion collector}, reveal that utilizing motion to specifically prompt feature $f^2_t$ yields superior performance compared to prompting either $f^3_t$ or $f^4_t$ individually or all features collectively. 
Fig.~\ref{fig:Prompt_destination_on_the_segmentation_stream} illustrates that applying a prompt on $f^2_t$  enables the model to focus on the most relevant object features while effectively suppressing background noise. However, when prompting all features, including $f^2_t$, mismatched prompt locations can introduce interference with the original features, ultimately degrading detection performance.
This finding suggests that semantic features at corresponding hierarchical levels between segmentation and motion tasks can synergistically enhance the representation of the original features. However, integrating features from different hierarchical levels, which lack proper alignment, may compromise the integrity and fidelity of the original feature representation, thereby negatively affecting overall performance.

\begin{table}[h!]
    \caption{Different prompt destinations on the segmentation stream.}
    \footnotesize
    \centering
    \tabcolsep=0.23cm
    \renewcommand{\arraystretch}{1.1}
    \begin{tabular}{ccc||ccccc}
    \toprule
      $f^2_t$ & $f^3_t$ & $f^4_t$ & $\mathcal{S}_\alpha\uparrow$ &$F_\beta^w\uparrow$ & $\mathcal{M}\downarrow$ & Dice$\uparrow$ & IoU$\uparrow$ \\
     \hline
    \Checkmark &  &  & \textbf{0.669} & \textbf{0.374} & \textbf{0.017} & \textbf{0.424} & \textbf{0.326} \\
     & \Checkmark &  & 0.642 & 0.319 & 0.019 & 0.359 & 0.271 \\
     &  & \Checkmark & 0.646 & 0.325 & 0.021 & 0.366 & 0.283 \\
    \Checkmark & \Checkmark & \Checkmark & 0.639 & 0.313 & 0.020 & 0.361 & 0.271 \\
    \bottomrule
    \end{tabular}
    \label{tab: posion of motion collector}
\end{table}

\subsubsection{Effectiveness of motion self-supervision} 
Quantitative and qualitative results are reported to validate the effectiveness of different training strategies for the motion stream. 
Due to the absence of optical flow ground truth, conducting quantitative analyses for the output optical flow is not feasible. Thus, we evaluate its effectiveness via segmentation/prediction performance. As validated in Table~\ref{tab: self-supervised loss}, the model with a self-supervised loss achieves better performance, which demonstrates that the prompted motion under self-supervision is beneficial to boosting detection performance. 

\begin{figure}[t!]
\begin{center}
\includegraphics[width=.481\textwidth]{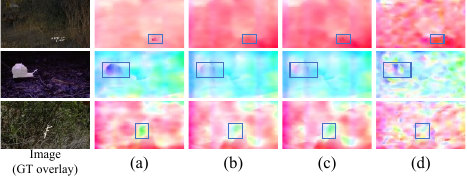}
\end{center}
\vspace{-8pt}
\caption{
Visual comparisons employing different designs:
(a) Our self-supervision (default EMIP),
(b) Full-tuning,
(c) GMFlow,
(d) Ours w/o self-supervised loss.}
\label{fig:Visualization_of}
\end{figure}

Fig. \ref{fig:Visualization_of} visualizes comparisons of our model with different settings in terms of optical flow prediction for camouflaged scenarios. 
The results show that sometimes the original GMFlow cannot perceive camouflaged moving objects (Fig. \ref{fig:Visualization_of} (c)), whereas ours that incorporates camouflage information with the prompt learning paradigm can better detect the targets (Fig. \ref{fig:Visualization_of} (a)). 
Fig. \ref{fig:Visualization_of} (d) further shows that the lack of self-supervised loss in our EMIP leads to worse flow prediction.
In contrast, full-tuning the GMFlow part (Fig. \ref{fig:Visualization_of} (b), correspond those results in Table~\ref{tab: frozen}) is hardly generalized in new camouflaged scenarios.

\begin{figure}[t!]
\begin{center}
\includegraphics[width=.492\textwidth]{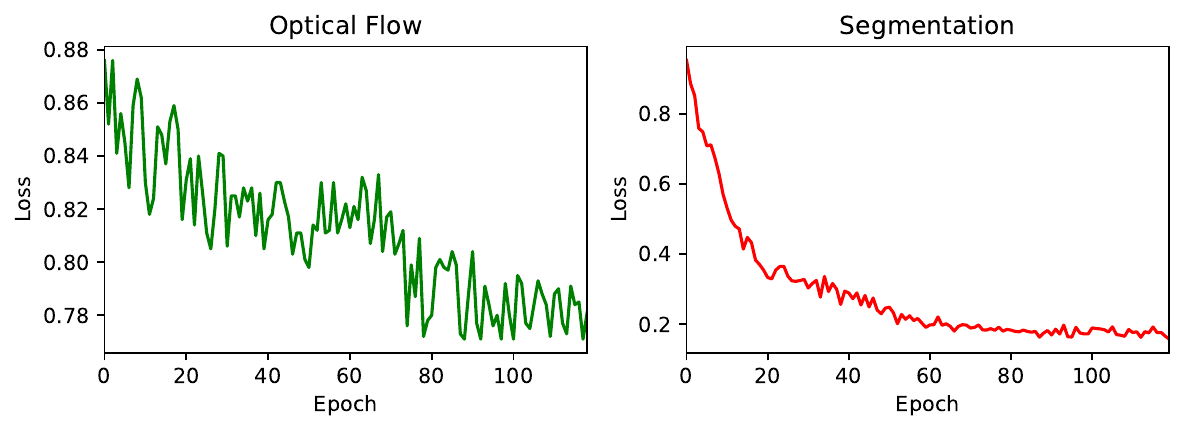}
\end{center}
\vspace{-8pt}
\caption{Training loss of optical flow and segmentation in our proposed model~\ourmodel. Both losses converge after about 80 epochs.}
\label{fig: Loss}
\end{figure}

Fig. \ref{fig: Loss} shows the loss curves of optical flow estimation (left) and camouflaged object segmentation (right), respectively, and one can see that both losses converge during the training process. Notably, the segmentation loss converges rapidly, while the flow loss converges at a slower rate. The straightforward utilization of a well pre-trained optical flow fundamental model results in an initially low flow loss. As the training proceeds, this flow loss enforces stable optimization of the segmentation-to-motion prompt learning, thereby guaranteeing a more effective prompt towards the motion stream.

\begin{table}[h!]
    \caption{Quantitative comparison of our EMIP model with and without the self-supervised loss for optical flow.}
    \footnotesize
    \centering
    \tabcolsep=0.25cm
    \renewcommand{\arraystretch}{1.1}
    \begin{tabular}{c||ccccc}
    \toprule
      Setting & $\mathcal{S}_\alpha\uparrow$ &$F_\beta^w\uparrow$ & $\mathcal{M}\downarrow$ & Dice$\uparrow$ & IoU$\uparrow$ \\
     \hline
     \textit{w/o} self-supervised & 0.644 & 0.321 & 0.020 & 0.363 & 0.277\\
     \textit{w/} self-supervised & \textbf{0.669} & \textbf{0.374} & \textbf{0.017} & \textbf{0.424} & \textbf{0.326} \\
    \bottomrule
    \end{tabular}
    \label{tab: self-supervised loss}
\end{table}

\subsection{Analyses of segmentation-to-motion prompt}
\begin{figure*}[t!]
    \centering
    \includegraphics[width=1.0\textwidth]{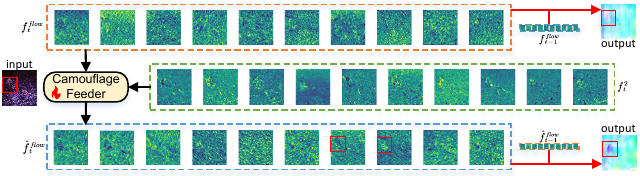}
    \vspace{-8pt}
    \caption{Visualization of the segmentation-to-motion prompt. The left image represents the current frame. The red rectangles highlights prominent regions in feature maps or the current frame. The solid red lines mean the computation processes of the output optical flow using motion features of two adjacent frames (\{$f_{t}^{flow}$,$f_{t-1}^{flow}$\} or \{$\hat{f}_{t}^{flow}$,$\hat{f}_{t-1}^{flow}$\}).}
    \label{fig: feature_map}
\end{figure*}

We introduce a segmentation-to-motion prompt strategy utilizing a camouflage feeder to enhance motion estimation. This approach refines the features that generate motion information, ensuring more precise responses in camouflaged regions while minimizing the impact of irrelevant information.
To better elucidate the inner mechanism and interpretability of segmentation-to-motion prompt, as illustrated in Fig. \ref{fig: feature_map}, we visualize several feature maps of $f^{flow}_t$, $\hat{f}^{flow}_t$, and $f^{2}_t$ in one frame. For a more intuitive presentation, we randomly select 10 consecutive feature maps along the channel dimension. As observed, the lower-level appearance features $f^{2}_t$ from the segmentation branch exhibits enhanced responses around the camouflaged regions.
After being prompted by appearance features, motion features then exhibit an enhanced response towards camouflaged regions. This response aids in further distinguishing the motion features of camouflaged areas from the background pixels, thereby improving the motion estimation for camouflaged objects.

\subsection{Computational Efficiency}
To thoroughly assess model-related parameters and efficiency, we perform a comparative analysis in Table \ref{tab: param} against the previous cutting-edge model SLT-Net~\cite{cheng2022implicit}, under the same GPU configuration. 
Despite having $\sim$18M more parameters compared to SLT-Net, EMIP exhibits decent improvements in both detection accuracy and frames-per-second (FPS).
Furthermore, the long-term variant of EMIP, \emph{i.e.}, EMIP$^\dag$, achieves such improvements with remarkably low amount of fine-tuned parameters (8.5M, less than 8\% of the total model parameters), reducing the computational overhead during training.

\begin{table}[h!]
    \caption{Comparison of model parameters and efficiency with previous SLT-Net, in terms of total parameters, finetuned parameters, and inference speed (frames-per-second, FPS). The best scores are highlighted in \textbf{bold}.}
    \footnotesize
    \centering
    \tabcolsep=0.13cm
    \renewcommand{\arraystretch}{1.1}
    \begin{tabular}{c||ccccccc}
    \toprule
      Model & Total Params & Finetuned Params & FPS & $\mathcal{S}_\alpha\uparrow$ &$F_\beta^w\uparrow$ & $\mathcal{M}\downarrow$\\
     \hline
     SLT-Net & \textbf{82.38M} & 82.38M & 5.5 & 0.637 & 0.304 & 0.027 \\
     EMIP & 100.86M & 96.06M & \textbf{7.8} & 0.669 & 0.374  & 0.017 \\
     EMIP$^\dag$ & 109.04M & \textbf{8.50M} & 6.2 & \textbf{0.675} & \textbf{0.381} & \textbf{0.015}\\
    \bottomrule
    \end{tabular}
    \label{tab: param}
\end{table}

\subsection{Generalization Ability on VSOD/VOS}
\begin{figure*}[t!]
    \centering
    \includegraphics[width=0.9\textwidth]{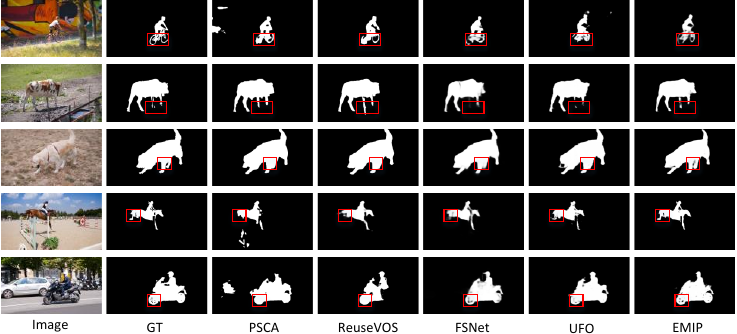}
    \vspace{-6pt}
    \caption{Visual comparisons of EMIP with four state-of-the-art VOS/VSOD methods. Red rectangles indicate challenging regions on which our EMIP excels.}
    \label{fig: generalization}
\end{figure*}

To thoroughly demonstrate the generalizability of \ourmodel, we conducted extended evaluations using the well-known VSOD/VOS datasets DAVIS\text{$_{16}$}~\cite{perazzi2016benchmark}, FBMS~\cite{ochs2013segmentation}, ViSal~\cite{wang2015consistent} and SegV2~\cite{li2013video}. 
The results, summarized in Table~\ref{tab: generalization}, include a comprehensive quantitative comparison against several recently published state-of-the-art methods for VSOD and VOS tasks. These methods represent a broad spectrum of current advancements in the field.
Additionally, we present visual comparisons in Fig. \ref{fig: generalization} to further illustrate the efficacy of our approach. Our model, \ourmodel, consistently excels in capturing fine contour details across various scenarios. For instance, it accurately delineates the foot of a dog in the 3rd row, the tail of a horse in the 4th row, and the wheel of a motorcycle in the 5th row. These detailed visual comparisons highlight our model's ability to handle intricate object boundaries and maintain high fidelity in the segmentation process.
The results clearly demonstrate that, although \ourmodel~is specifically designed to tackle video camouflage scenes, it also performs exceptionally well in more general video segmentation tasks. This adaptability underscores the robustness and versatility of our approach, making it suitable for a wide range of applications beyond its original design scope.

\begin{table*}[t]
\caption{
Comparisons of our \ourmodel~with other state-of-the-art VSOD and VOS methods on VSOD datasets. The majority of the results are borrowed from~\cite{zhao2024motion} or acquired from their released code weights. Unavailable metrics are denoted by -.
$\dag$ denotes video segmentation methods trained on DAVIS17~\cite{pont20172017} and YouTube-VOS~\cite{xu2018youtube} datasets, whose results are acquired from their released code weights.
The best results are \textbf{bolded} for highlighting.
}
\label{tab: generalization}
    \renewcommand\arraystretch{1.1}
	\begin{center}
		\scalebox{0.98}{
			\begin{tabular}{c|ccc|ccc|ccc|ccc}
				\toprule  
				\multirow{2}*{Method} &
				\multicolumn{3}{c|}{DAVIS\text{$_{16}$}} & \multicolumn{3}{c|}{FBMS} &  \multicolumn{3}{c|}{ViSal} &
				\multicolumn{3}{c}{SegV2}
				\\
				\cline{2-13}
	        {~}&
				\textit{$\mathcal{S}_\alpha\uparrow$} & \textit{$\mathcal{F}\uparrow$} & $\mathcal{M}\downarrow$ &
			 \textit{$\mathcal{S}_\alpha\uparrow$} & \textit{$\mathcal{F}\uparrow$} & $\mathcal{M}\downarrow$ &
				\textit{$\mathcal{S}_\alpha\uparrow$} & \textit{$\mathcal{F}\uparrow$} & $\mathcal{M}\downarrow$  & \textit{$\mathcal{S}_\alpha\uparrow$} & \textit{$\mathcal{F}\uparrow$}  & 
				$\mathcal{M}\downarrow$\\
				\toprule  
				
				SCOM\cite{chen2018scom}\tiny{$_{\text{TIP'2018}}$}  & 0.832  & 0.783 &  0.048  & 0.794 & 0.797 & 0.079  & 0.762 & 0.831 & 0.122  & 0.815 & 0.764 & 0.030\\
				
				MBNM\cite{li2018unsupervised}\tiny{$_{\text{ECCV'2018}}$}  & 0.887  & 0.861 &  0.031  & 0.857 & 0.816 & 0.047  & 0.898 & 0.883 & 0.020  & 0.809 & 0.716 & 0.026\\
				
				PDBM\cite{song2018pyramid}\tiny{$_{\text{ECCV'2018}}$}  & 0.882 & 0.855 &  0.028  & 0.851  & 0.821 & 0.064  & 0.907  & 0.888 & 0.032  & 0.864 & 0.800 & 0.024 \\

				SRP\cite{cong2019video}\tiny{$_{\text{TIP'2019}}$}  & 0.662  & 0.660 & 0.070   & 0.648  & 0.671 & 0.134  & - & 0.752 & 0.092 & - & 0.683 & 0.095 \\
				
				MESO\cite{xu2019video}\tiny{$_{\text{TMM'2019}}$}  & 0.718  & 0.660 & 0.070  & 0.635  & 0.618 & 0.134 & - & -  & - & - & - & -\\
				
				LTSI\cite{chen2019improved}\tiny{$_{\text{TIP'2019}}$} & 0.876  & 0.850 & 0.034  & 0.805  & 0.799 & 0.087  & 0.922 & 0.909 & 0.027  & 0.827 & 0.862 & 0.028 \\
				
				RSE\cite{xu2019video1}\tiny{$_{\text{TCSVT'2019}}$} & 0.748  & 0.698 & 0.063  & 0.670  & 0.652 & 0.128 & - & - & - & - & - & -\\
				
				SSAV\cite{fan2019shifting}\tiny{$_{\text{CVPR'2019}}$} & 0.893  & 0.861 & 0.028  & {0.879} & 0.865 & {0.040}  & 0.943 & 0.939 & 0.020  & 0.851 & 0.801 & 0.023 \\
				
				RCR\cite{yan2019semi}\tiny{$_{\text{ICCV'2019}}$} & 0.886 & 0.848 & 0.027   & 0.872  & 0.859  & 0.053 & - & -  & - & - & - & -\\

				CAS\cite{ji2020casnet}\tiny{$_{\text{TNNLS'2020}}$} & 0.873 & 0.860 & 0.032  & 0.856 & 0.863 & 0.056  & - & -  & - & 0.820 & 0.847 & 0.029 \\

				PCSA\cite{gu2020pyramid}\tiny{$_{\text{AAAI'2020}}$} & 0.902 &  0.880 & 0.022  & 0.868  & 0.837 & 0.040 & 0.946 & 0.940  & 0.017  & 0.865 & 0.810 & 0.025 \\
				
				DFNet\cite{zhen2020learning}\tiny{$_{\text{ECCV'2020}}$} & - & 0.899 & 0.018 & - & 0.833 & 0.054  & - & 0.927 & 0.017  & - & - & -\\
			
				ReuseVOS$\dag$\cite{park2021learning}\tiny{$_{\text{CVPR'2021}}$}& 0.883  & 0.865 & 0.019  & 0.888 & 0.884 & \textbf{0.027}  & 0.928 & 0.933 & 0.020  & 0.844 & 0.832 & 0.025 \\
				
				TransVOS$\dag$\cite{mei2021transvos}\tiny{$_{\text{PrePrint'2021}}$}& 0.885  & 0.869 & 0.018  & 0.867 & 0.886 & 0.038  & 0.917 & 0.928 & 0.021  & 0.816 & 0.800 & 0.024 \\

                UFO~\cite{su2023unified}\tiny{$_{\text{TMM'2023}}$} & 0.874 & 0.797 & 0.032  & 0.868 & 0.803 & 0.041  & 0.940 & 0.914 & 0.012  & 0.836 & 0.746 & 0.057 \\
                
                MAMNet~\cite{zhao2024motion}\tiny{$_{\text{TIP'2024}}$}& 0.897 & 0.877  & 0.020  & \textbf{0.894} & 0.883 & 0.032  & 0.947 & 0.948 & 0.012  & 0.886 & 0.850 & 0.014 \\
					
				\midrule  
				\textbf{EMIP \textit{(Ours)}} & \textbf{0.908}  & \textbf{0.902} & \textbf{0.016}  & 0.891 & \textbf{0.887} & 0.032  & \textbf{0.950}  & \textbf{0.950} & \textbf{0.012}  & \textbf{0.891} & \textbf{0.862} & \textbf{0.013} \\			

				\bottomrule 
		\end{tabular}}    
\end{center}
\vspace{-8pt}
\end{table*}

\subsection{Failure Cases}
\begin{figure}[t!]
\begin{center}
\includegraphics[width=.481\textwidth]{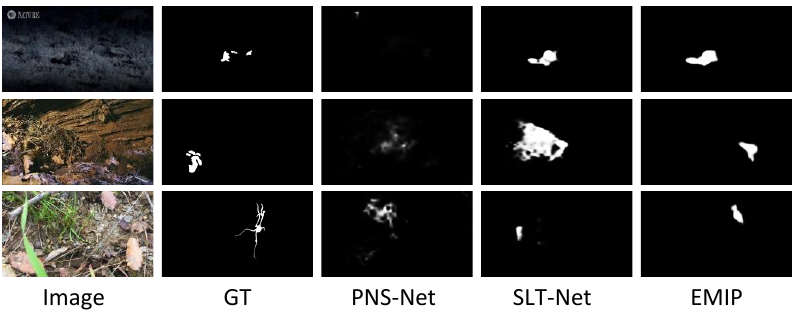}
\end{center}
\vspace{-8pt}
\caption{Some failure cases of EMIP and two most recent methods.}
\label{fig: failure cases}
\end{figure}

In the 1st scenario depicted in  Fig.~\ref{fig: failure cases}, the camouflaged object is occluded. In this context, the detection outcome encompasses the occlusion object due to the significant resemblance across the camouflaged object, occlusion entity, and background.
Another scenario is shown in the 2nd row of the figure. The displayed image represents a frame among initial frames of a sequence, lacking motion information and posing a challenge even for human visual perception, let alone learning models. 
For the 3rd scenario, the prominent green leaves in the foreground act as strong distractors and the camouflaged object is easily mistaken as part of the foreground foliage.
It is important to note that these challenges are not unique to our model. Even state-of-the-art methods from previous research, such as PNS-Net~\cite{ji2021progressively} and SLT-Net~\cite{cheng2022implicit}, encounter similar difficulties, as shown in Fig.~\ref{fig: failure cases}. Consequently, addressing these specific challenges remains a crucial direction for future research in the field of VCOD.

\section{Conclusion}
We propose EMIP, an innovative framework for VCOD that explicitly handles motion cues through the utilization of a frozen pre-trained optical flow fundamental model. 
EMIP adopts a novel two-stream architecture, concurrently addressing camouflaged segmentation and optical flow estimation. 
The core idea of interaction between these two streams is orchestrated through an interactive prompting mechanism.  
Experimental results show that the paradigm of interactive prompting of EMIP can enhance the outputs of both streams, further achieving accurate prediction. 
Comprehensive ablation studies and in-depth discussions validate the key components of EMIP.
In addition, we present an extended version of EMIP, incorporating historical features into the prompt to alleviate short-term prediction errors and enhance overall accuracy.
Moreover, the proposed EMIP is extended to the general video object segmentation task, consistently delivering improved performance and validating its generalizability and adaptability.
Our contributions not only achieve compelling results on two VCOD benchmark datasets, but also provide fresh insights into addressing the challenging VCOD task. 
We hope that the proposed framework could serve as a catalyst for inspiring further research in this emerging field.
We believe that making controllable and adjustable optimization prompts for fundamental models presents an intriguing avenue for future investigation.




\bibliographystyle{IEEEtran}
\bibliography{main.bbl}

\begin{thebibliography}{10}
\providecommand{\url}[1]{#1}
\csname url@samestyle\endcsname
\providecommand{\newblock}{\relax}
\providecommand{\bibinfo}[2]{#2}
\providecommand{\BIBentrySTDinterwordspacing}{\spaceskip=0pt\relax}
\providecommand{\BIBentryALTinterwordstretchfactor}{4}
\providecommand{\BIBentryALTinterwordspacing}{\spaceskip=\fontdimen2\font plus
\BIBentryALTinterwordstretchfactor\fontdimen3\font minus \fontdimen4\font\relax}
\providecommand{\BIBforeignlanguage}[2]{{%
\expandafter\ifx\csname l@#1\endcsname\relax
\typeout{** WARNING: IEEEtran.bst: No hyphenation pattern has been}%
\typeout{** loaded for the language `#1'. Using the pattern for}%
\typeout{** the default language instead.}%
\else
\language=\csname l@#1\endcsname
\fi
#2}}
\providecommand{\BIBdecl}{\relax}
\BIBdecl

\bibitem{zhao2019object}
Z.-Q. Zhao, P.~Zheng, S.-t. Xu, and X.~Wu, ``Object detection with deep learning: A review,'' \emph{IEEE TNNLS}, vol.~30, no.~11, pp. 3212--3232, 2019.

\bibitem{zhuge2022salient}
M.~Zhuge, D.-P. Fan, N.~Liu, D.~Zhang, D.~Xu, and L.~Shao, ``Salient object detection via integrity learning,'' \emph{IEEE TPAMI}, vol.~45, no.~3, pp. 3738--3752, 2023.

\bibitem{ji2022vps}
G.-P. Ji, G.~Xiao, Y.-C. Chou, D.-P. Fan, K.~Zhao, G.~Chen, and L.~Van~Gool, ``Video polyp segmentation: A deep learning perspective,'' \emph{Machine Intelligence Research}, vol.~19, no.~6, pp. 531--549, 2022.

\bibitem{wu2023medical}
J.~Wu, R.~Fu, H.~Fang, Y.~Liu, Z.~Wang, Y.~Xu, Y.~Jin, and T.~Arbel, ``Medical sam adapter: Adapting segment anything model for medical image segmentation,'' \emph{arXiv preprint arXiv:2304.12620}, 2023.

\bibitem{he2019fully}
T.~He, Y.~Liu, C.~Xu, X.~Zhou, Z.~Hu, and J.~Fan, ``A fully convolutional neural network for wood defect location and identification,'' \emph{IEEE Access}, vol.~7, pp. 123\,453--123\,462, 2019.

\bibitem{yang2022surface}
K.~Yang, Y.~Liu, S.~Zhang, and J.~Cao, ``Surface defect detection of heat sink based on lightweight fully convolutional network,'' \emph{IEEE TIM}, vol.~71, pp. 1--12, 2022.

\bibitem{fan2021concealed}
D.-P. Fan, G.-P. Ji, M.-M. Cheng, and L.~Shao, ``Concealed object detection,'' \emph{IEEE TPAMI}, 2021.

\bibitem{fan2020Camouflage}
D.-P. Fan, G.-P. Ji, G.~Sun, M.-M. Cheng, J.~Shen, and L.~Shao, ``Camouflaged object detection,'' in \emph{CVPR}, 2020.

\bibitem{ji2023gradient}
G.-P. Ji, D.-P. Fan, Y.-C. Chou, D.~Dai, A.~Liniger, and L.~Van~Gool, ``Deep gradient learning for efficient camouflaged object detection,'' \emph{Machine Intelligence Research}, vol.~20, no.~1, pp. 92--108, 2023.

\bibitem{cheng2022implicit}
X.~Cheng, H.~Xiong, D.-P. Fan, Y.~Zhong, M.~Harandi, T.~Drummond, and Z.~Ge, ``Implicit motion handling for video camouflaged object detection,'' in \emph{CVPR}, 2022.

\bibitem{lamdouar2020betrayed}
H.~Lamdouar, C.~Yang, W.~Xie, and A.~Zisserman, ``Betrayed by motion: Camouflaged object discovery via motion segmentation,'' in \emph{ACCV}, 2020.

\bibitem{xie2022segmenting}
J.~Xie, W.~Xie, and A.~Zisserman, ``Segmenting moving objects via an object-centric layered representation,'' in \emph{NeurIPS}, 2022.

\bibitem{yang2021selfsupervised}
C.~Yang, H.~Lamdouar, E.~Lu, A.~Zisserman, and W.~Xie, ``Self-supervised video object segmentation by motion grouping,'' in \emph{ICCV}, 2021.

\bibitem{teed2020raft}
Z.~Teed and J.~Deng, ``Raft: Recurrent all-pairs field transforms for optical flow,'' in \emph{ECCV}.\hskip 1em plus 0.5em minus 0.4em\relax Springer, 2020, pp. 402--419.

\bibitem{liu2020learning}
L.~Liu, J.~Zhang, R.~He, Y.~Liu, Y.~Wang, Y.~Tai, D.~Luo, C.~Wang, J.~Li, and F.~Huang, ``Learning by analogy: Reliable supervision from transformations for unsupervised optical flow estimation,'' in \emph{CVPR}, 2020, pp. 6489--6498.

\bibitem{xu2022gmflow}
H.~Xu, J.~Zhang, J.~Cai, H.~Rezatofighi, and D.~Tao, ``Gmflow: Learning optical flow via global matching,'' in \emph{CVPR}, 2022, pp. 8121--8130.

\bibitem{sandler2022fine}
M.~Sandler, A.~Zhmoginov, M.~Vladymyrov, and A.~Jackson, ``Fine-tuning image transformers using learnable memory,'' in \emph{CVPR}, 2022, pp. 12\,155--12\,164.

\bibitem{jia2022visual}
M.~Jia, L.~Tang, B.-C. Chen, C.~Cardie, S.~Belongie, B.~Hariharan, and S.-N. Lim, ``Visual prompt tuning,'' in \emph{ECCV}.\hskip 1em plus 0.5em minus 0.4em\relax Springer, 2022, pp. 709--727.

\bibitem{zhu2023visual}
J.~Zhu, S.~Lai, X.~Chen, D.~Wang, and H.~Lu, ``Visual prompt multi-modal tracking,'' in \emph{CVPR}, 2023, pp. 9516--9526.

\bibitem{pan2011study}
Y.~Pan, Y.~Chen, Q.~Fu, P.~Zhang, X.~Xu \emph{et~al.}, ``Study on the camouflaged target detection method based on 3d convexity,'' \emph{Modern Applied Science}, vol.~5, no.~4, p. 152, 2011.

\bibitem{liu2012foreground}
Z.~Liu, K.~Huang, and T.~Tan, ``Foreground object detection using top-down information based on em framework,'' \emph{IEEE TIP}, vol.~21, no.~9, pp. 4204--4217, 2012.

\bibitem{Li2017TGWV}
S.~Li, D.~Florencio, Y.~Zhao, C.~Cook, and W.~Li, ``Foreground detection in camouflaged scenes,'' in \emph{ICIP}, 2017, pp. 4247--4251.

\bibitem{Li2018fusion}
S.~Li, D.~Florencio, W.~Li, Y.~Zhao, and C.~Cook, ``A fusion framework for camouflaged moving foreground detection in the wavelet domain,'' \emph{IEEE TIP}, vol.~27, no.~8, pp. 3918--3930, 2018.

\bibitem{garcia2020background}
B.~Garcia-Garcia, T.~Bouwmans, and A.~J.~R. Silva, ``Background subtraction in real applications: Challenges, current models and future directions,'' \emph{Computer Science Review}, vol.~35, p. 100204, 2020.

\bibitem{mei2021camouflaged}
H.~Mei, G.-P. Ji, Z.~Wei, X.~Yang, X.~Wei, and D.-P. Fan, ``Camouflaged object segmentation with distraction mining,'' in \emph{CVPR}, 2021, pp. 8772--8781.

\bibitem{zhai2021mutual}
Q.~Zhai, X.~Li, F.~Yang, C.~Chen, H.~Cheng, and D.-P. Fan, ``Mutual graph learning for camouflaged object detection,'' in \emph{CVPR}, 2021, pp. 12\,997--13\,007.

\bibitem{liu2021POCINet}
Y.~Liu, D.~Zhang, Q.~Zhang, and J.~Han, ``Integrating part-object relationship and contrast for camouflaged object detection,'' \emph{IEEE TIFS}, vol.~16, pp. 5154--5166, 2021.

\bibitem{pang2022zoom}
Y.~Pang, X.~Zhao, T.-Z. Xiang, L.~Zhang, and H.~Lu, ``Zoom in and out: A mixed-scale triplet network for camouflaged object detection,'' in \emph{CVPR}, 2022, pp. 2160--2170.

\bibitem{pang2024zoomnext}
------, ``Zoomnext: A unified collaborative pyramid network for camouflaged object detection,'' \emph{IEEE TPAMI}, vol.~46, no.~12, pp. 9205--9220, 2024.

\bibitem{jia2022segmar}
Q.~Jia, S.~Yao, Y.~Liu, X.~Fan, R.~Liu, and Z.~Luo, ``Segment, magnify and reiterate: Detecting camouflaged objects the hard way,'' in \emph{CVPR}, 2022, pp. 4713--4722.

\bibitem{huang2023feature}
Z.~Huang, H.~Dai, T.-Z. Xiang, S.~Wang, H.-X. Chen, J.~Qin, and H.~Xiong, ``Feature shrinkage pyramid for camouflaged object detection with transformers,'' in \emph{CVPR}, 2023, pp. 5557--5566.

\bibitem{zhang2023predictive}
Y.~Zhang, J.~Zhang, W.~Hamidouche, and O.~Deforges, ``Predictive uncertainty estimation for camouflaged object detection,'' \emph{IEEE TIP}, 2023.

\bibitem{hu2023high}
X.~Hu, S.~Wang, X.~Qin, H.~Dai, W.~Ren, D.~Luo, Y.~Tai, and L.~Shao, ``High-resolution iterative feedback network for camouflaged object detection,'' in \emph{AAAI}, vol.~37, no.~1, 2023, pp. 881--889.

\bibitem{he2023FEDER}
C.~He, K.~Li, Y.~Zhang, L.~Tang, Y.~Zhang, Z.~Guo, and X.~Li, ``Camouflaged object detection with feature decomposition and edge reconstruction,'' in \emph{CVPR}, 2023, pp. 22\,046--22\,055.

\bibitem{cong2023frequency}
R.~Cong, M.~Sun, S.~Zhang, X.~Zhou, W.~Zhang, and Y.~Zhao, ``Frequency perception network for camouflaged object detection,'' in \emph{ACM MM}, 2023, pp. 1179--1189.

\bibitem{Yao2024HGINet}
S.~Yao, H.~Sun, T.-Z. Xiang, X.~Wang, and X.~Cao, ``Hierarchical graph interaction transformer with dynamic token clustering for camouflaged object detection,'' \emph{IEEE TIP}, vol.~33, pp. 5936--5948, 2024.

\bibitem{hao2025senet}
C.~Hao, Z.~Yu, X.~Liu, J.~Xu, H.~Yue, and J.~Yang, ``A simple yet effective network based on vision transformer for camouflaged object and salient object detection,'' \emph{IEEE TIP}, vol.~34, pp. 608--622, 2025.

\bibitem{bideau2016s}
P.~Bideau and E.~Learned-Miller, ``It’s moving! a probabilistic model for causal motion segmentation in moving camera videos,'' in \emph{ECCV}, 2016, pp. 433--449.

\bibitem{zhang2017CM}
X.~Zhang, C.~Zhu, S.~Wang, Y.~Liu, and M.~Ye, ``A bayesian approach to camouflaged moving object detection,'' \emph{IEEE TCSVT}, vol.~27, no.~9, pp. 2001--2013, 2017.

\bibitem{hui2024implict-explicit}
W.~Hui, Z.~Zhu, G.~Gu, M.~Liu, and Y.~Zhao, ``Implicit-explicit motion learning for video camouflaged object detection,'' \emph{IEEE TMM}, pp. 1--9, 2024.

\bibitem{lu2025weakly-supervised}
Z.~Lu, L.~Xie, X.~Zhao, B.~Xu, H.~Liang, and R.~Liang, ``A weakly-supervised cross-domain query framework for video camouflage object detection,'' \emph{IEEE TCSVT}, vol.~35, no.~2, pp. 1506--1518, 2025.

\bibitem{hui2024endow}
W.~Hui, Z.~Zhu, S.~Zheng, and Y.~Zhao, ``Endow sam with keen eyes: Temporal-spatial prompt learning for video camouflaged object detection,'' in \emph{CVPR}, 2024, pp. 19\,058--19\,067.

\bibitem{mei2021transvos}
J.~Mei, M.~Wang, Y.~Lin, Y.~Yuan, and Y.~Liu, ``Transvos: Video object segmentation with transformers,'' \emph{arXiv preprint arXiv:2106.00588}, 2021.

\bibitem{park2021learning}
H.~Park, J.~Yoo, S.~Jeong, G.~Venkatesh, and N.~Kwak, ``Learning dynamic network using a reuse gate function in semi-supervised video object segmentation,'' in \emph{CVPR}, 2021, pp. 8405--8414.

\bibitem{li2022recurrent}
M.~Li, L.~Hu, Z.~Xiong, B.~Zhang, P.~Pan, and D.~Liu, ``Recurrent dynamic embedding for video object segmentation,'' in \emph{CVPR}, 2022, pp. 1332--1341.

\bibitem{oh2019video}
S.~W. Oh, J.-Y. Lee, N.~Xu, and S.~J. Kim, ``Video object segmentation using space-time memory networks,'' in \emph{ICCV}, 2019, pp. 9226--9235.

\bibitem{chen2018scom}
Y.~Chen, W.~Zou, Y.~Tang, X.~Li, C.~Xu, and N.~Komodakis, ``Scom: Spatiotemporal constrained optimization for salient object detection,'' \emph{IEEE TIP}, vol.~27, no.~7, pp. 3345--3357, 2018.

\bibitem{li2018unsupervised}
S.~Li, B.~Seybold, A.~Vorobyov, X.~Lei, and C.-C.~J. Kuo, ``Unsupervised video object segmentation with motion-based bilateral networks,'' in \emph{ECCV}, 2018, pp. 207--223.

\bibitem{song2018pyramid}
H.~Song, W.~Wang, S.~Zhao, J.~Shen, and K.-M. Lam, ``Pyramid dilated deeper convlstm for video salient object detection,'' in \emph{ECCV}, 2018, pp. 715--731.

\bibitem{cong2019video}
R.~Cong, J.~Lei, H.~Fu, F.~Porikli, Q.~Huang, and C.~Hou, ``Video saliency detection via sparsity-based reconstruction and propagation,'' \emph{IEEE TIP}, vol.~28, no.~10, pp. 4819--4831, 2019.

\bibitem{xu2019video}
M.~Xu, B.~Liu, P.~Fu, J.~Li, and Y.~H. Hu, ``Video saliency detection via graph clustering with motion energy and spatiotemporal objectness,'' \emph{IEEE TMM}, vol.~21, no.~11, pp. 2790--2805, 2019.

\bibitem{yan2019semi}
P.~Yan, G.~Li, Y.~Xie, Z.~Li, C.~Wang, T.~Chen, and L.~Lin, ``Semi-supervised video salient object detection using pseudo-labels,'' in \emph{ICCV}, 2019.

\bibitem{wang2018non}
X.~Wang, R.~Girshick, A.~Gupta, and K.~He, ``Non-local neural networks,'' in \emph{CVPR}, 2018, pp. 7794--7803.

\bibitem{ballas2016delving}
N.~Ballas, L.~Yao, C.~J. Pal, and A.~Courville, ``Delving deeper into convolutional networks for learning video representations,'' in \emph{ICLR}, 2016.

\bibitem{chen2021exploring}
C.~Chen, G.~Wang, C.~Peng, Y.~Fang, D.~Zhang, and H.~Qin, ``Exploring rich and efficient spatial temporal interactions for real-time video salient object detection,'' \emph{IEEE TIP}, vol.~30, pp. 3995--4007, 2021.

\bibitem{ji2021full}
G.-P. Ji, K.~Fu, Z.~Wu, D.-P. Fan, J.~Shen, and L.~Shao, ``Full-duplex strategy for video object segmentation,'' in \emph{ICCV}, 2021, pp. 4922--4933.

\bibitem{zhao2024motion}
X.~Zhao, H.~Liang, P.~Li, G.~Sun, D.~Zhao, R.~Liang, and X.~He, ``Motion-aware memory network for fast video salient object detection,'' \emph{IEEE TIP}, 2024.

\bibitem{brown2020language}
T.~Brown, B.~Mann, N.~Ryder, M.~Subbiah, J.~D. Kaplan, P.~Dhariwal, A.~Neelakantan, P.~Shyam, G.~Sastry, A.~Askell \emph{et~al.}, ``Language models are few-shot learners,'' in \emph{NeurIPS}, vol.~33, 2020, pp. 1877--1901.

\bibitem{kirillov2023segment}
A.~Kirillov, E.~Mintun, N.~Ravi, H.~Mao, C.~Rolland, L.~Gustafson, T.~Xiao, S.~Whitehead, A.~C. Berg, W.-Y. Lo \emph{et~al.}, ``Segment anything,'' in \emph{ICCV}, 2023, pp. 4015--4026.

\bibitem{wang2022pvt}
W.~Wang, E.~Xie, X.~Li, D.-P. Fan, K.~Song, D.~Liang, T.~Lu, P.~Luo, and L.~Shao, ``Pvt v2: Improved baselines with pyramid vision transformer,'' \emph{Computational Visual Media}, vol.~8, no.~3, pp. 415--424, 2022.

\bibitem{fan2020pra}
D.-P. Fan, G.-P. Ji, T.~Zhou, G.~Chen, H.~Fu, J.~Shen, and L.~Shao, ``Pranet: Parallel reverse attention network for polyp segmentation,'' in \emph{MICCAI}, 2020.

\bibitem{zeiler2014visualizing}
M.~D. Zeiler and R.~Fergus, ``Visualizing and understanding convolutional networks,'' in \emph{ECCV}.\hskip 1em plus 0.5em minus 0.4em\relax Springer, 2014, pp. 818--833.

\bibitem{zamir2022restormer}
S.~W. Zamir, A.~Arora, S.~Khan, M.~Hayat, F.~S. Khan, and M.-H. Yang, ``Restormer: Efficient transformer for high-resolution image restoration,'' in \emph{CVPR}, 2022, pp. 5728--5739.

\bibitem{hendrycks2016gaussian}
D.~Hendrycks and K.~Gimpel, ``Gaussian error linear units (gelus),'' \emph{arXiv preprint arXiv:1606.08415}, 2016.

\bibitem{vaswani2017attention}
A.~Vaswani, N.~Shazeer, N.~Parmar, J.~Uszkoreit, L.~Jones, A.~N. Gomez, {\L}.~Kaiser, and I.~Polosukhin, ``Attention is all you need,'' in \emph{NeurIPS}, vol.~30, 2017.

\bibitem{liu2021amd}
R.~Liu, Z.~Wu, S.~Yu, and S.~Lin, ``The emergence of objectness: Learning zero-shot segmentation from videos,'' in \emph{NeurIPS}, vol.~34, 2021, pp. 13\,137--13\,152.

\bibitem{ji2023sam}
G.-P. Ji, D.-P. Fan, P.~Xu, M.-M. Cheng, B.~Zhou, and L.~Van~Gool, ``Sam struggles in concealed scenes--empirical study on" segment anything",'' \emph{Science China Information Sciences}, 2023.

\bibitem{wang2004image}
Z.~Wang, A.~C. Bovik, H.~R. Sheikh, and E.~P. Simoncelli, ``Image quality assessment: from error visibility to structural similarity,'' \emph{IEEE TIP}, vol.~13, no.~4, pp. 600--612, 2004.

\bibitem{21Fan_HybridLoss}
D.-P. Fan, G.-P. Ji, X.~Qin, and M.-M. Cheng, ``Cognitive vision inspired object segmentation metric and loss function,'' \emph{Scientia Sinica Informationis}, vol.~6, no.~6, p.~5, 2021.

\bibitem{perazzi2016benchmark}
F.~Perazzi, J.~Pont-Tuset, B.~McWilliams, L.~Van~Gool, M.~Gross, and A.~Sorkine-Hornung, ``A benchmark dataset and evaluation methodology for video object segmentation,'' in \emph{CVPR}, 2016, pp. 724--732.

\bibitem{ochs2013segmentation}
P.~Ochs, J.~Malik, and T.~Brox, ``Segmentation of moving objects by long term video analysis,'' \emph{IEEE TPAMI}, vol.~36, no.~6, pp. 1187--1200, 2013.

\bibitem{wang2015consistent}
W.~Wang, J.~Shen, and L.~Shao, ``Consistent video saliency using local gradient flow optimization and global refinement,'' \emph{IEEE TIP}, vol.~24, no.~11, pp. 4185--4196, 2015.

\bibitem{li2013video}
F.~Li, T.~Kim, A.~Humayun, D.~Tsai, and J.~M. Rehg, ``Video segmentation by tracking many figure-ground segments,'' in \emph{ICCV}, 2013, pp. 2192--2199.

\bibitem{fan2017structure}
D.-P. Fan, M.-M. Cheng, Y.~Liu, T.~Li, and A.~Borji, ``{Structure-measure: A New Way to Evaluate Foreground Maps},'' in \emph{ICCV}, 2017.

\bibitem{margolin2014evaluate}
R.~Margolin, L.~Zelnik-Manor, and A.~Tal, ``How to evaluate foreground maps?'' in \emph{CVPR}, 2014.

\bibitem{achanta2009frequency}
R.~Achanta, S.~Hemami, F.~Estrada, and S.~Susstrunk, ``Frequency-tuned salient region detection,'' in \emph{CVPR}, 2009, pp. 1597--1604.

\bibitem{perazzi2012saliency}
F.~Perazzi, P.~Kr{\"a}henb{\"u}hl, Y.~Pritch, and A.~Hornung, ``Saliency filters: Contrast based filtering for salient region detection,'' in \emph{CVPR}, 2012, pp. 733--740.

\bibitem{kingma2014adam}
D.~P. Kingma and J.~Ba, ``Adam: A method for stochastic optimization,'' \emph{arXiv preprint arXiv:1412.6980}, 2014.

\bibitem{paszke2019pytorch}
A.~Paszke, S.~Gross, F.~Massa, A.~Lerer, J.~Bradbury, G.~Chanan, T.~Killeen, Z.~Lin, N.~Gimelshein, L.~Antiga \emph{et~al.}, ``Pytorch: An imperative style, high-performance deep learning library,'' in \emph{NeurIPS}, vol.~32, 2019.

\bibitem{zhao2019EGNet}
J.-X. Zhao, J.-J. Liu, D.-P. Fan, Y.~Cao, J.~Yang, and M.-M. Cheng, ``Egnet: Edge guidance network for salient object detection,'' in \emph{ICCV}, 2019.

\bibitem{Qin_2019_CVPR}
X.~Qin, Z.~Zhang, C.~Huang, C.~Gao, M.~Dehghan, and M.~Jagersand, ``Basnet: Boundary-aware salient object detection,'' in \emph{CVPR}, 2019.

\bibitem{Wu_2019_CVPR}
Z.~Wu, L.~Su, and Q.~Huang, ``Cascaded partial decoder for fast and accurate salient object detection,'' in \emph{CVPR}, 2019.

\bibitem{sun2024frequency}
Y.~Sun, C.~Xu, J.~Yang, H.~Xuan, and L.~Luo, ``Frequency-spatial entanglement learning for camouflaged object detection,'' in \emph{ECCV}.\hskip 1em plus 0.5em minus 0.4em\relax Springer, 2024, pp. 343--360.

\bibitem{ji2021progressively}
G.-P. Ji, Y.-C. Chou, D.-P. Fan, G.~Chen, H.~Fu, D.~Jha, and L.~Shao, ``Progressively normalized self-attention network for video polyp segmentation,'' in \emph{MICCAI}.\hskip 1em plus 0.5em minus 0.4em\relax Springer, 2021, pp. 142--152.

\bibitem{pont20172017}
J.~Pont-Tuset, F.~Perazzi, S.~Caelles, P.~Arbel{\'a}ez, A.~Sorkine-Hornung, and L.~Van~Gool, ``The 2017 davis challenge on video object segmentation,'' \emph{arXiv preprint arXiv:1704.00675}, 2017.

\bibitem{xu2018youtube}
N.~Xu, L.~Yang, Y.~Fan, D.~Yue, Y.~Liang, J.~Yang, and T.~Huang, ``Youtube-vos: A large-scale video object segmentation benchmark,'' \emph{arXiv preprint arXiv:1809.03327}, 2018.

\bibitem{chen2019improved}
C.~Chen, G.~Wang, C.~Peng, X.~Zhang, and H.~Qin, ``Improved robust video saliency detection based on long-term spatial-temporal information,'' \emph{IEEE TIP}, vol.~29, pp. 1090--1100, 2019.

\bibitem{xu2019video1}
M.~Xu, B.~Liu, P.~Fu, J.~Li, Y.~H. Hu, and S.~Feng, ``Video salient object detection via robust seeds extraction and multi-graphs manifold propagation,'' \emph{IEEE TCSVT}, vol.~30, no.~7, pp. 2191--2206, 2019.

\bibitem{fan2019shifting}
D.-P. Fan, W.~Wang, M.-M. Cheng, and J.~Shen, ``Shifting more attention to video salient object detection,'' in \emph{CVPR}, 2019, pp. 8554--8564.

\bibitem{ji2020casnet}
Y.~Ji, H.~Zhang, Z.~Jie, L.~Ma, and Q.~J. Wu, ``Casnet: A cross-attention siamese network for video salient object detection,'' \emph{IEEE TNNLS}, vol.~32, no.~6, pp. 2676--2690, 2020.

\bibitem{gu2020pyramid}
Y.~Gu, L.~Wang, Z.~Wang, Y.~Liu, M.-M. Cheng, and S.-P. Lu, ``Pyramid constrained self-attention network for fast video salient object detection,'' in \emph{AAAI}, vol.~34, no.~07, 2020, pp. 10\,869--10\,876.

\bibitem{zhen2020learning}
M.~Zhen, S.~Li, L.~Zhou, J.~Shang, H.~Feng, T.~Fang, and L.~Quan, ``Learning discriminative feature with crf for unsupervised video object segmentation,'' in \emph{ECCV}.\hskip 1em plus 0.5em minus 0.4em\relax Springer, 2020, pp. 445--462.

\bibitem{su2023unified}
Y.~Su, J.~Deng, R.~Sun, G.~Lin, H.~Su, and Q.~Wu, ``A unified transformer framework for group-based segmentation: Co-segmentation, co-saliency detection and video salient object detection,'' \emph{IEEE TMM}, 2023.

\end{thebibliography}


 



\vfill

\end{document}